\pdfoutput=1
\documentclass{article}

    \PassOptionsToPackage{numbers, compress}{natbib}
\usepackage[preprint]{neurips_2024}
\usepackage{threeparttable}
\usepackage{tcolorbox}
\usepackage{enumitem}
\usepackage{soul}
\usepackage[pagebackref=true,breaklinks=true,colorlinks,citecolor=citecolor, linkcolor=linkcolor]{hyperref}

\usepackage[T1]{fontenc}    % use 8-bit T1 fonts
\usepackage{hyperref}       % hyperlinks
\usepackage{url}            % simple URL typesetting
\usepackage{booktabs}       % professional-quality tables
\usepackage{amsfonts}       % blackboard math symbols
\usepackage{nicefrac}       % compact symbols for 1/2, etc.
\usepackage{microtype}      % microtypography
\usepackage{xcolor, colortbl}         % colors
\usepackage{graphicx}
\usepackage{amsmath}
\usepackage{algorithm}
\usepackage{algorithmic}
\usepackage{multirow}
\usepackage{array}
\definecolor{citecolor}{HTML}{0071BC}
\definecolor{linkcolor}{HTML}{ED1C24}
\usepackage{tcolorbox}

\title{HuatuoGPT-Vision, Towards Injecting Medical Visual Knowledge into Multimodal LLMs at Scale}

\definecolor{lightpink}{rgb}{0.945, 0.816, 0.804}
\definecolor{lightgreen}{rgb}{0.851, 0.906, 0.839}
\definecolor{lightblue}{rgb}{0.8, 0.9, 1}
\definecolor{lightyellow}{rgb}{0.992, 0.949, 0.816}
\newcommand{\dataset}{\textbf{PubMedVision}}
\newcommand{\imp}[1]{{\small\hspace{0.05cm}{\color[HTML]{32CB00}{$_{\textbf{+#1}}$}}}}

\author{Junying Chen$^{1,2}$, 
Chi Gui$^{2}$, 
Ruyi Ouyang$^{2}$, Anningzhe Gao$^{1,2}$,
 Shunian Chen$^{1,2}$ \\ \textbf{Guiming Hardy Chen}$^{1,2}$,  \textbf{Xidong Wang}$^{1,2}$, \textbf{Ruifei Zhang}$^{1,2}$, \textbf{Zhenyang Cai}$^{1,2}$,
\textbf{Ke Ji}$^{1,2}$\\  \textbf{Guangjun Yu}$^{1,2,3}$, \textbf{Xiang Wan}$^{1,2,3}$, \textbf{Benyou Wang}$^{1,2}$\thanks{Benyou is the corresponding author with email: \textit{wangbenyou@cuhk.edu.cn}.} \\
$^1$ Shenzhen Research Institute of Big Data \\
$^2$ The Chinese University of Hong Kong, Shenzhen \\
$^3$ National Health Data Institute, Shenzhen\\
\url{https://github.com/FreedomIntelligence/HuatuoGPT-Vision}\\
\url{https://huggingface.co/datasets/FreedomIntelligence/PubMedVision}\\
}

\begin{document}

\maketitle

\begin{abstract}

The rapid development of multimodal large language models (MLLMs), such as GPT-4V, has led to significant advancements. However, these models still face challenges in medical multimodal capabilities due to limitations in the quantity and quality of medical vision-text data, stemming from data privacy concerns and high annotation costs. While pioneering approaches utilize PubMed's large-scale, de-identified medical image-text pairs to address these limitations, they still fall short due to inherent data noise. To tackle this, we refined medical image-text pairs from PubMed and employed MLLMs (GPT-4V) in an \textit{'unblinded'} capacity to denoise and reformat the data, resulting in the creation of the \textbf{PubMedVision} dataset with 1.3 million medical VQA samples. Our validation demonstrates that: \textbf{(1)} PubMedVision can significantly enhance the medical multimodal capabilities of current MLLMs, showing significant improvement in benchmarks including the MMMU Health \& Medicine track; \textbf{(2)} manual checks by medical experts and empirical results validate the superior data quality of our dataset compared to other data construction methods. Using PubMedVision, we train a 34B medical MLLM \textbf{HuatuoGPT-Vision}, which shows superior performance in medical multimodal scenarios among open-source MLLMs. 
% Our code and data are available at \url{https://github.com/FreedomIntelligence/HuatuoGPT-Vision} and \url{https://huggingface.co/datasets/FreedomIntelligence/PubMedVision}.

\end{abstract}

\section{Introduction}

Multimodal Large Language Models (MLLMs), such as GPT4-V, show limited performance in medical applications, particularly in lacking visual knowledge specific to the medical domain~\cite{gpt4v-fail1,gpt4v-fail2}. Although there are some small-scale, high-quality datasets containing medical visual knowledge~\cite{rad-vqa,slake,pathvqa}, scaling them up is challenging. Additionally, there are privacy and licensing issues associated with medical data, further complicating matters.

Pioneering works \cite{pmc-vqa, llava-med-citation, RadFM} utilize PubMed\footnote{PubMed is a free search engine that primarily accesses the MEDLINE database, containing references and scientific papers on life sciences and biomedical topics.} for larger-scale training for medical vision-language alignment. PubMed is favored because it contains medical images and surrounding text, which (i) encapsulate the forefront of human wisdom in medicine and (ii) are well-de-identified \cite{pmc-oa}. However, models trained on PubMed are unsatisfactory, as they perform poorly compared to general MLLMs on medical multimodal tasks~\citep{omnimedvqa,generalmorewell2}. This can be attributed to data noise in PubMed, which significantly affects multimodal performance~\citep{llava, allava}.

Concurrently, LLaVA-Med \cite{llava-med-citation} uses a ``\textit{blind}'' Large Language Model (LLM) to generate Visual Question Answering (VQA) from the contextual text of PubMed images, achieving notable results. However, this approach might overlook visual information inherent in the medical images themselves as LLMs cannot perceive images as input, probably leading to the generation of misinterpreted or irrelevant answers. Moreover, LLaVA-Med is limited to 56K medical VQA entries. Thus, creating a higher-quality and larger-scale 
vision-language alignment dataset for medicine is essential.

To close this gap, we meticulously select high-quality medical image-text pair from PubMed, employing a proposed refined pipeline. Utilizing \textit{914,960} refined medical images and their corresponding text, we apply GPT-4V as the ``\textit{unblinded}'' reformatter, contrasting the ``\textit{blinded}'' reformatting used in previous works~\cite{llava-med-citation,RadFM,pmc-vqa}, to denoise the PubMed data. Our method generates more aligned medical VQA data for medical multimodal  alignment. Consequently, we constructed a high-quality multimodal medical dataset with \textit{1.3 million} samples and name it as \dataset.

Our experiments validated \dataset\ in two key aspects: (1) It significantly enhances the medical multimodal capabilities of MLLMs, showing notable improvement in benchmarks such as MMMU Health \& Medicine. LLaVA-v1.5-LLaMA-3-8B achieves the strongest performance among open-source MLLMs with PubMedVision ; (2)  Manual checks by medical experts and empirical results confirmed the superior data quality of PubMedVision compared to current data construction methods.
% \dataset\ is available at \url{https://huggingface.co/datasets/FreedomIntelligence/PubMedVision}.

The contributions of this paper are summarized as follows:
\begin{enumerate}
\item  \textbf{Unblinded Data Reformatting for Medical Multimodality.} We propose leveraging ``\textit{unblinded}'' MLLMs to reformat PubMed image-text pairs to construct a better-aligned medical VQA dataset. Expert reviews and empirical tests show that this method yields higher-quality data, improving MLLM training.

% \item \textbf{PubMedVision: A Large-scale, High-quality Medical Multimodal Dataset.}  Using the ``\textit{unblinded}'' reformatter, we bulid \dataset, containing 1.3 million VQA samples from PubMed medical images and contextual text. Experiments demonstrate that PubMedVision significantly enhances MLLMs' medical multimodal capabilities, enabling models like \textit{LLaVA-1.5-LLaMA-3-8B} to outperform other general and medical open-source MLLMs.
\item \textbf{PubMedVision: A Large-scale, High-quality Medical Multimodal Dataset.} With the MLLM-powered reformatted method, we bulid \dataset, containing 1.3 million medical VQA entries for visual alignment. Experiments demonstrate that PubMedVision significantly enhances MLLMs' medical multimodal capabilities, enabling models like \textit{LLaVA-1.5-LLaMA-3-8B} to outperform other general and medical open-source MLLMs.

\item \textbf{HuatuoGPT-Vision: A Medical MLLM.} Using PubMedVision, we trained \textbf{HuatuoGPT-Vision}, a 34B parameter medical MLLM. HuatuoGPT-Vision demonstrate superior performance on multiple medical multimodal benchmarks among open-source models.

\end{enumerate}

\section{Medical Visual Alignment in  MLLMs }

\subsection{Existing Medical VQA Data}

Table \ref{tb-vqa} compares existing medical VQA datasets, which are crucial for image-text alignment and instruction following in medical MLLMs. Early datasets like VQA-RAD, SLAKE, and Path-VQA are limited by their small size (less than 20K entries) and their exclusive focus on radiology. PMC-CaseReport, PMC-VQA, and LLaVA-Med leverage PubMed medical images to scale data and employ LLMs to reformat contextual text into VQA. However, these datasets also suffer from limited quantity and are prone to misinterpretation and misalignment due to the 'blinded' nature of the LLMs.  In contrast, we aim to construct a larger-scale, high-quality medical VQA dataset, PubMedVision.

\begin{table}[ht!]
\centering\small
\begin{tabular}{lcccccc} \toprule
    & Data Size & Modality   & \begin{tabular}[c]{@{}c@{}}Uses\\LLMs?\end{tabular} & \begin{tabular}[c]{@{}c@{}}Uses\\MLLMs?\end{tabular} & Source  \\ \midrule
VQA-RAD \cite{rad-vqa}       & 3,515       & Radiology & $\times$            & $\times$    &      Websites  \\
SLAKE   \cite{slake}       & 14,028      & Radiology   & $\times$            & $\times$     &       Websites \\
PathVQA  \cite{pathvqa}     & 17,325     & Pathology    & $\times$            & $\times$       &     Books \\
% RP3D-VQA      & 60,685      & Radiology  & $\checkmark$  & $\checkmark$        & $\times$             \\
PMC-CaseReport  \cite{RadFM}    & 54,341      & Radiology    & $\checkmark$        & $\times$ & PubMed            \\
PMC-VQA \cite{pmc-vqa}      & 176,919     & Multimodal   & $\checkmark$        & $\times$   & PubMed      \\
LLaVA-Med VQA \cite{llava-med-citation} & 56,702      & Multimodal  & $\checkmark$        & $\times$  & PubMed           \\
\textbf{\colorbox{lightblue}{PubMedVision (Ours)}}      & 1,294,062   & Multimodal  & $\times$            & $\checkmark$  & PubMed      \\ \bottomrule
\end{tabular}
\caption{\label{tb-vqa}Comparison of Medical VQA Datasets }
\end{table}

\subsection{Medical Visual Alignment through the Lens of  Data  Engineering}
\label{Section:2_2}

\paragraph{Visual Knowledge Alignment} Current MLLMs typically adapt a text-only LLM with a visual encoder \cite{llava, li2023blip}. Therefore, \textit{alignment} involves injecting image knowledge into LLMs, aligning images with the language understanding of LLMs. This paper explores the injection of extensive medical visual knowledge from PubMed into MLLMs, as PubMed is a leading repository of advanced medical research with well-de-identified medical images.

\textbf{Data Noises in PubMed}
Although existing work~\cite{RadFM,llava-med-citation,pmc-vqa} utilize PubMed, it has not been entirely satisfactory, as they still lag behind many general-purpose MLLMs in medical vision \cite{omnimedvqa,generalmorewell2}. We attribute it to the data noises in PubMed. The text surrounding the image in PubMed papers does not always well-describe the image. While relevant, this text does not necessarily facilitate effective visual alignment.

\textbf{The Efforts to Improve Data Quality Sourced from PubMed}
The original data is not always suitable for training, as seen in reformatting alignment~\cite{fan2024reformatted}. Compared to \colorbox{lightpink}{Native Captions} in PubMed, existing work uses text-only LLMs to reformat these captions of images, denoted as \colorbox{lightgreen}{LLM-Reformatted}. This can result in misinterpreted or misaligned text for the images due to the \textit{blined} LLM. To solve this, we propose using a \textit{multimodal} LLM, called \colorbox{lightblue}{MLLM-Reformatted}.  Additionally, we compare with \colorbox{lightyellow}{GPT4v-Distill}, a popular approach to distill GPT-4V in general multimodal fields, such as ShareGPT4V \cite{sharegpt4v} and ALLaVA-4V \cite{allava}. For GPT4v-Distilled, we provide only images to GPT-4V to generate a medical description.

\begin{figure*}[ht!]
  \centering
  \resizebox{0.95\textwidth}{!}{
  \includegraphics[width=\textwidth]{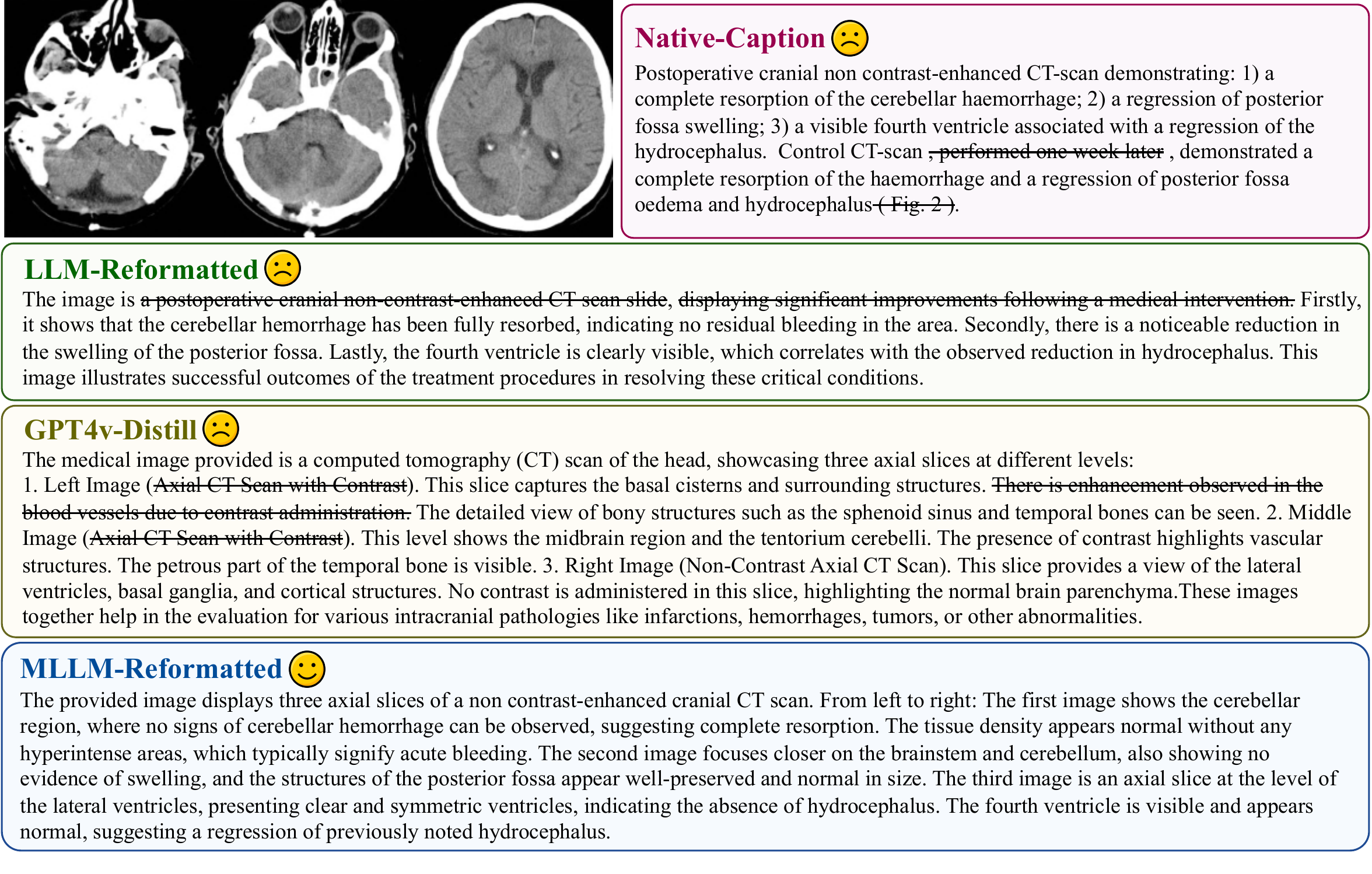}
  }
  % \caption{\label{fig-case} Constructing image captions in various approaches. We provide detailed explanations of these four methods in Appendix \ref{ap-methoddetail}. We use \textit{gpt-4-turbo-2024-04-09} as the LLM and \textit{gpt-4-turbo-2024-04-09} as the MLLM and GPT4-V. 
  % \st{strikethrough  texts} indicate some hallucinated info probably because the reformatting LLM  cannot perceive the images.  Contrastively, with MLLM for reformatting it could accurately describes the image's main content and key information.
  % This case is source from PubMed paper at \url{https://www.ncbi.nlm.nih.gov/pmc/articles/PMC2852039/}.}
  \caption{\label{fig-case} Constructing image captions in various approaches. Detailed explanations of these methods are given in Appendix \ref{ap-methoddetail}. We use \textit{gpt-4} as the LLM and \textit{gpt-4V} as the MLLM. \st{Strikethrough texts} indicate erroneous descriptions or descriptions unrelated to the image. This case is sourced from a PubMed paper at \url{https://www.ncbi.nlm.nih.gov/pmc/articles/PMC2852039/}.}
\end{figure*}

\vspace{-5mm}
\paragraph{Case Analysis} Figure \ref{fig-case} presents examples generated by these methods. It can be observed that \colorbox{lightpink}{Native-Caption} captions are ambiguous and contain content unrelated to the image. \colorbox{lightgreen}{LLM-Reformatted} misinterprets three sub-images as a CT slide, leading to misleading descriptions, and fails to exclude irrelevant content. \colorbox{lightyellow}{GPT4v-Distill} generates factually incorrect descriptions due to the lack of contextual text. In contrast, \colorbox{lightblue}{MLLM-Reformatted} produces superior descriptions by leveraging both visual information and contextual cues. It accurately and thoroughly describes the key information of the image. The subsequent experiment in Section \ref{section:ep2} further demonstrates the higher data quality of MLLM-Reformatted.

\section{PubMedVision}
\begin{figure*}[ht!]
  \centering
  \resizebox{1\textwidth}{!}{
  \includegraphics[width=\textwidth]{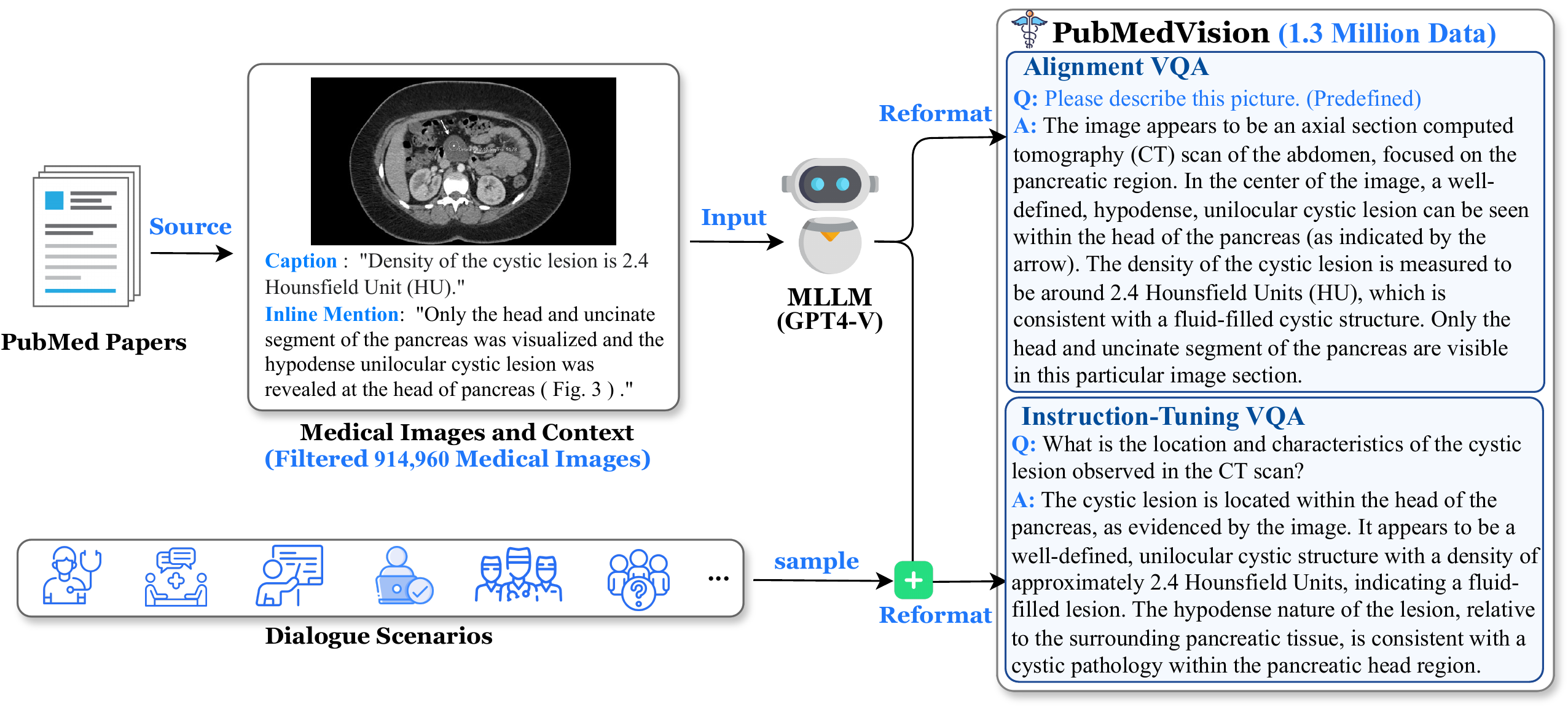}
  }
  \caption{\label{fig-f1}Construction process of the PubMedVision dataset.}
\end{figure*}
% The data process of PubMedVision is shown in Figure \ref{fig-f1}.

\subsection{Data Collection}

To acquire a comprehensive dataset of PubMed medical images, we integrated previously compiled public data of PubMed images, specifically LLaVA-Med PMC (514K) \cite{llava-med-citation}, PMC-Inline (11M) \cite{RadFM}, and PMC-OA (1M) \cite{pmc-oa}. Although extensive, the majority of this merged data consists of charts and graphs from papers rather than medical images. Therefore, we implemented a rigorous data filtering pipeline:  \textbf{(1) Text Filtering.} A medical vocabulary was used to filter out data where the contextal text contains a insufficient number of medical terms.  \textbf{(2) Image Filtering.} We excluded low-resolution images (less than 336x336 pixels). A medical image classification model, trained on 1K manually labeled images and 10K MLLM-labeled images, is used to identify medical images. \textbf{(3) Deduplication.} Using Sentence-BERT \cite{reimers2019sentence} as the encoder, we obtained semantic embeddings of the image captions and filtered out images with overly similar contexts. For more details, please see Appendix \ref{ap-datapipline}.

\begin{figure}[ht!] 
    \centering
    \begin{minipage}{0.5\textwidth}
        \centering
  \includegraphics[width=\textwidth]{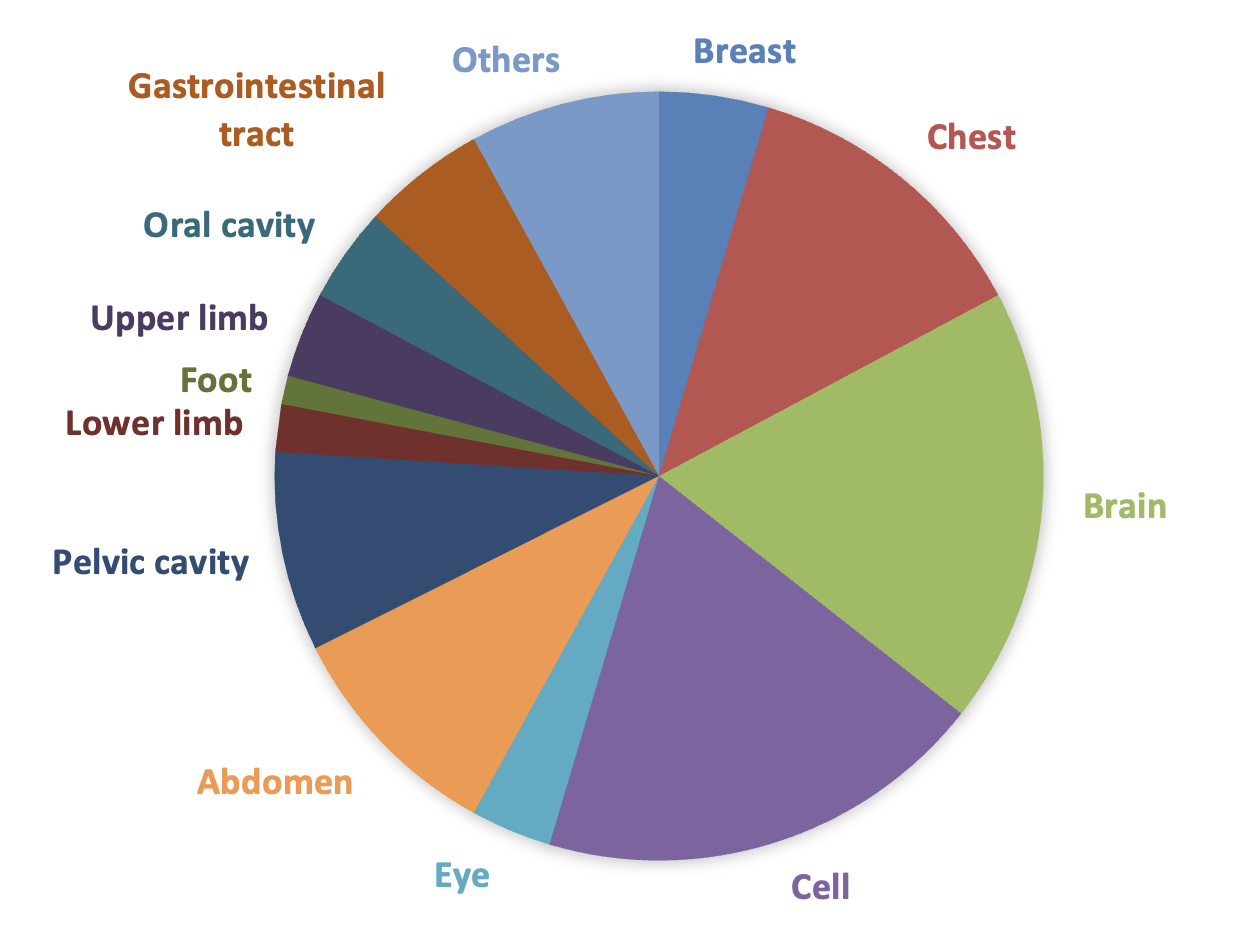}
    \end{minipage}\hfill
    \begin{minipage}{0.45\textwidth}
         \centering
  \includegraphics[width=\textwidth]{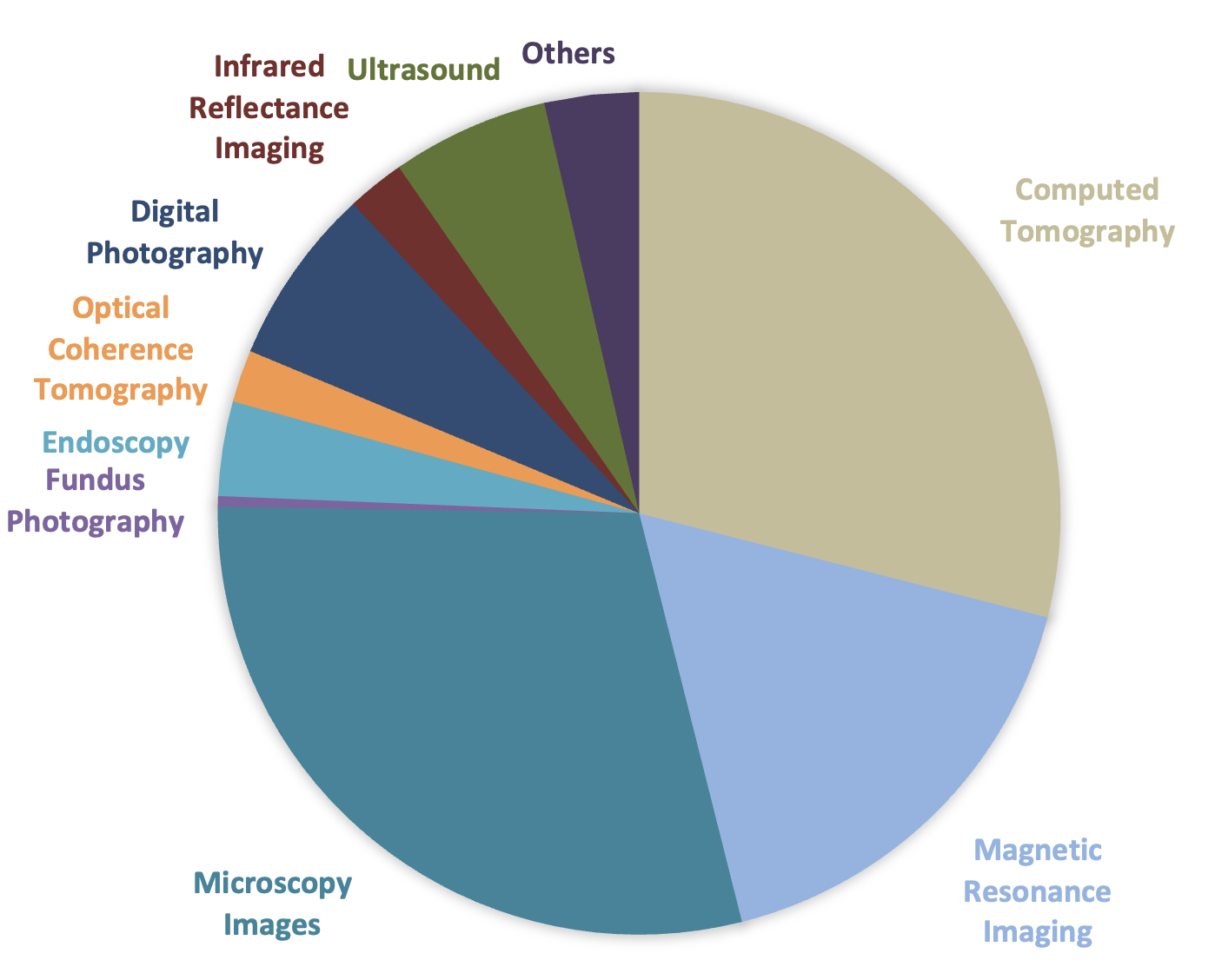}
    \end{minipage}
    \caption{\label{fig-datatype}Image Diversity in PubMedVision. A random sample of 500 images from PubMedVision are categorized. \textbf{Left:} Distribution of body parts depicted in the images. \textbf{Right:} Distribution of imaging modalities.}
\end{figure}

Ultimately, we filtered out 914,960 medical images and their associated contextual text (captions and inline mentions). Figure \ref{fig-datatype} illustrates the diversity of medical modalities and image regions covered by PubMedVision's images. These medical images are then used to sequentially construct 1.3 million VQA data points for medical alignment.

\definecolor{outerboxcolor}{gray}{0.90} % 外部方框的颜色
\subsection{Data Reformatting with MLLMs}
\begin{figure*}[ht!]
  \centering
  \resizebox{1\textwidth}{!}{
  \includegraphics[width=\textwidth]{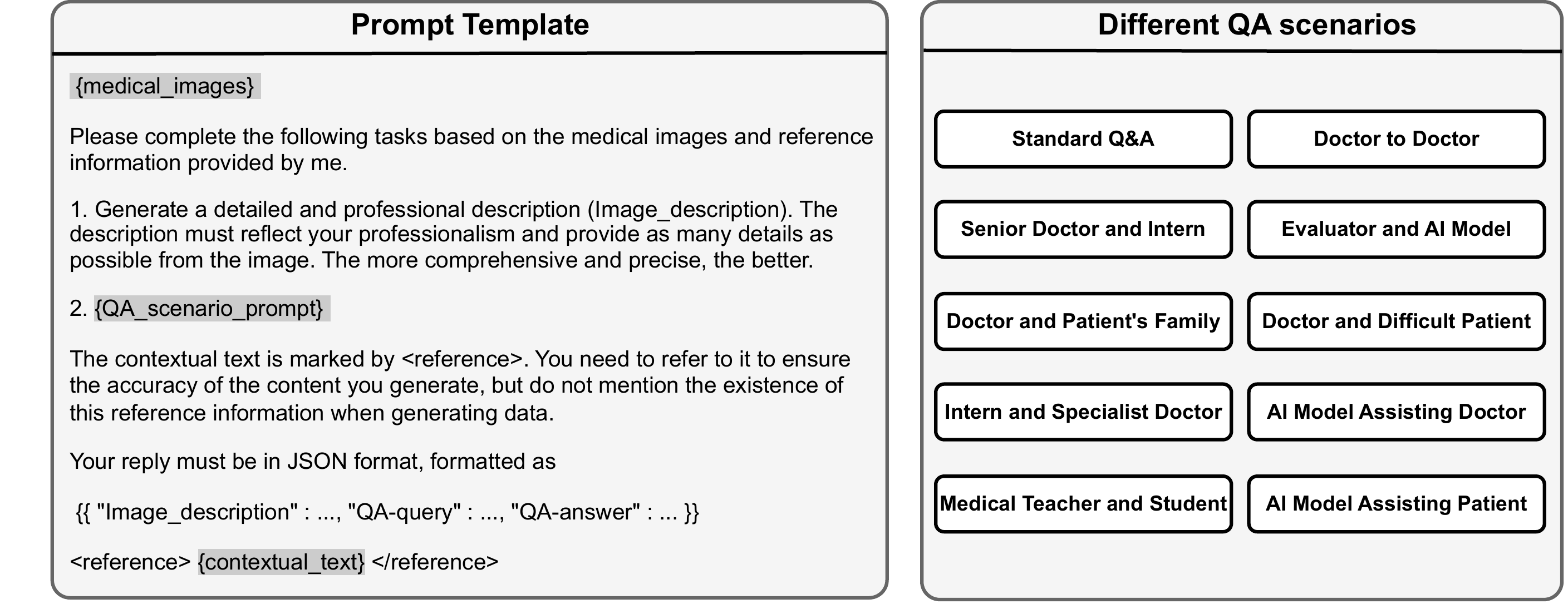}
  }
  \caption{\label{fig-prompt} Prompts used for data generation. \textbf{\{medical\_images\}} represents medical images. \textbf{\{QA\_scenario\_prompt\}} denotes scenario prompts, sampled from the scenarios on the right, see Appendix \ref{ap-ScenariosPrompt} for details. \textbf{\{contextual\_text\}} pertains to image captions and inline mentions.}
\end{figure*}

Each collected data point includes one or more medical images $\mathcal{I}$ and their corresponding contextual image descriptions $X$. As shown in Figure \ref{fig-f1}, we provided $\mathcal{I}$ and $X$ to MLLMs to generate medical VQA data. According to ALLaVA \cite{allava}, we generate two types of VQA data to enhance image alignment. Using the prompt shown in Figure \ref{fig-prompt}, the MLLM generates an overall image description $d$, a specific question $q$ about the image, and the corresponding answer $a$, as follows:
$$d, q, a = \mathrm{MLLMs}(\mathcal{I}, X)$$

\paragraph{Alignment VQA} We predefined a question $q'$ and combined it with the image description $d$ to form Alignment VQA $(q',a)$. The predefined question was sampled from a set of predefined questions, which can be found in Appendix \ref{ap-questionset}. According to ShareGPT-4V \cite{sharegpt4v}, such detailed image descriptions help in learning the alignment from image to text. 
% The Alignment VQA could be used in the first stage of training MLLMs \cite{llava}.

\paragraph{Instruction-Tuning VQA} We used the question $q$ and answer $a$ generated by MLLMs as Instruction-Tuning VQA $(q,a)$ for enhancing instruction-following ability and image comprehension. Unlike Alignment VQA, the questions are generated by MLLMs specifically for the images. To diversify the generated $q$, we designed eight different scenarios, as detailed in Appendix \ref{ap-ScenariosPrompt}. We randomly sample scenario settings into the synthetic prompt to enable MLLMs to generate more varied questions.

Based on this method, we employ GPT-4V~(\textit{gpt-4-turbo-2024-04-09}) as MLLMs to synthesize 647,031 Alignment VQA and 647,031 Instruction-Tuning VQA. Consequently, PubMedVision contains a total of 1.3 million data points.

\section{Experiment}
\subsection{Experiment Settings}

\paragraph{Training and Validation} To verify the effectiveness of PubMedVision, we selected the LLaVA-1.5 model architecture combined with LLaMA-3-8B. We use the original settings of LLaVA-1.5, featuring a 336×336 CLIP-Large mode \cite{clip} and a two-layer MLP Projector. For the base LLM, we utilize LLaMA-3-8B, which is pre-trained on OpenHermes \cite{OpenHermes} text instruction data. We followed the same two-stage training method as LLaVA-1.5 \cite{llava} (Pretraining and Finetuning) and the same hyperparameters (including a learning rate of 2e-5 and one epoch). Based on this setup, we train the following three comparative models:

\begin{itemize}
\item \textbf{LLaVA-v1.5-LLaMA3-8B}  The baseline model that only uses LLaVA-1.5 data. The data distribution is Pretraining: \textbf{558K} (LLaVA); Finetuning: \textbf{658K} (LLaVA).

\item \textbf{LLaVA-v1.5-LLaMA3-8B + LLaVA\_Med} This model uses both LLaVA-1.5 data and LLaVA\_Med's two-stage data. The data distribution is Pretraining: \textbf{558K} (LLaVA) + \textbf{ 457K} (LLaVA\_Med Alignment); Finetuning: \textbf{658K} (LLaVA) + \textbf{57K} (LLaVA\_Med VQA).

\item \textbf{LLaVA-v1.5-LLaMA3-8B + \colorbox{lightblue}{PubMedVision}} This model uses both LLaVA-1.5 data and PubMedVision data. The data distribution is Pretraining: \textbf{558K} (LLaVA) + \textbf{647K} (PubMedVision Alignment VQA); Finetuning: \textbf{658K} (LLaVA) + \textbf{647K} (PubMedVision Instruction-Tuning VQA).
\end{itemize}

\paragraph{HuatuoGPT-Vision} Building on PubMedVision, we developed our specialized medical MLLM, HuatuoGPT-Vision. It enhances \textit{LLaVA-v1.5-LLaMA3-8B + \colorbox{lightblue}{PubMedVision}} by featuring: (1) a larger model, utilizing Yi-1.5-34B \cite{yi} as the foundational LLM; (2) bilingual capabilities, supported by an additional 348K \textit{Chinese} medical VQA dataset translated from PubMedVision; and (3) enhanced medical knowledge, with added training from the medical text corpus of HuatuoGPT-II \cite{chen2023huatuogpt}.

\paragraph{Baselines} We compared two types of open-source models: \textbf{(1) Medical MLLMs.} We evaluated three Medical MLLMs, including Med-Flamingo \cite{Med-flamingo}, RadFM \cite{RadFM}, and LLaVA-Med-7B \cite{llava-med-citation}. \textbf{(2) General MLLMs.} We compared the latest models in the LLaVA series, including LLaVA-v1.6-7B, LLaVA-v1.6-13B, and LLaVA-v1.6-34B \cite{llava-1.6}. Additionally, we included comparisons with Yi-VL-34B \cite{yi} and Qwen-VL-Chat \cite{qwen}. 

\paragraph{Benchmarks}
To verify the medical multimodal capabilities of MLLMs, we employed three types of benchmarks:
\textbf{(1) Medical VQA Benchmark,} for which we used the test sets of VQA-RAD \cite{rad-vqa}, SLAKE \cite{slake}, PathVQA \cite{pathvqa}, and PMC-VQA \cite{pmc-vqa} to assess medical question-answering capabilities. Specifically, for SLAKE, we evaluated using its English CLOSED segment. 
\textbf{(2) Multimodal Benchmark: } MMMU \cite{mmmu} is a popular multimodal benchmark, and we utilized the Health \& Medicine track of MMMU, which is relevant to medical multimodality.
\textbf{(3) Traditional Medical Imaging Tasks.} We used the open access part of the OmniMedVQA dataset \cite{omnimedvqa}, which includes 42 traditional medical imaging datasets, all formatted as VQA.
{Note that for all benchmarks, we use the \textbf{zero-shot} method and the question template set by LLaVA, as shown in Appendix \ref{ap-evaluationTemplate}. }

% \paragraph{Open Source Baselines}

\subsection{Experiment 1: Effectiveness of PubMedVision}

\begin{table}[ht!]
\centering \small
\begin{tabular}{lcccc|c}
\toprule 
          \textbf{Model}                        & \textbf{VQA-RAD} & \textbf{SLAKE} & \textbf{PathVQA} & \textbf{PMC-VQA} & \textbf{Avg.} \\ \midrule
Med-Flamingo                      & 45.4    & 43.5  & 54.7    & 23.3  & 41.7 \\
RadFM                             & 50.6    & 34.6  & 38.7    & 25.9   & 37.5 \\
LLaVA-Med-7B                         & 51.4    & 48.6  & 56.8    & 24.7  & 45.4 \\
Qwen-VL-Chat  & 47.0    & 56.0  & 55.1    & 36.6  & 48.9 \\
Yi-VL-34B  &  53.0    & 58.9  & 47.3    & 39.5 & 49.7   \\
LLaVA-v1.6-7B              & 52.6    & 57.9  & 47.9    & 35.5  & 48.5 \\
LLaVA-v1.6-13B             & 55.8    & 58.9  & 51.9    & 36.6  & 50.8 \\
LLaVA-v1.6-34B             & 58.6    & 67.3  & 59.1    & 44.4  & 57.4 \\ 
\rowcolor{gray!17}\multicolumn{6}{l}{\textbf{Our Training}}                                     \\
LLaVA-v1.5-LLaMA3-8B              & 54.2    & 59.4  & 54.1    & 36.4  & 51.0  \\
\quad + LLaVA\_Med & 60.2    & 61.2  & 54.5   & 46.6   & 55.6  \\
\quad +  \colorbox{lightblue}{PubMedVision}  & \textbf{63.8}    & \textbf{74.5}  & \textbf{59.9}    & \textbf{52.7}  & \textbf{62.7} \\ \midrule
\textbf{HuatuoGPT-Vision-34B}  & \textbf{68.1}    & \textbf{76.9}  & \textbf{63.5}    & \textbf{58.2}  & \textbf{66.7} \\ 
\bottomrule
\end{tabular} 
\caption{\label{tb-res1} The results of the medical VQA benchmark.}
\end{table}

\paragraph{Medical VQA Benchmarks} Table \ref{tb-res1} presents the results of the medical VQA benchmarks. General-purpose MLLMs, such as LLaVA-v1.6, demonstrate superior performance compared to medical-specific MLLMs like LLaVA-Med-7B, aligning with the findings of prior studies \cite{omnimedvqa}.
However, the addition of medical multimodal data to LLaVA-v1.5-LLaMA3-8B significantly enhances performance, revealing substantial potential for improving medical image understanding. Notably, the use of the PubMedVision led to an \textbf{11.7\%} increase in overall accuracy, significantly outperforming the earlier LLaVA\_Med dataset. Additionally, as detailed in Appendix \ref{ap-moreexperiment}, fine-tuning on the training sets of these four datasets indicates that PubMedVision can also significantly improves performance in downstream medical multimodal tasks.

\begin{table}[ht!]
\centering\small
\begin{tabular}{lcccccccc|c}
\toprule 
\textbf{Model}   &   \textbf{CT}   &   \textbf{FP}   &   \textbf{MRI}   &   \textbf{OCT}   &   \textbf{Der}   &   \textbf{Mic}   &   \textbf{X-Ray}   &   \textbf{US}   &   \textbf{Avg.}   \\ \midrule
Med-Flamingo  & 34.6& 33.3& 27.5& 26.0& 28.3& 28.1& 30.1& 33.2& 30.2 \\
RadFM & 33.3& 35.0& 22.0& 31.3& 36.3& 28.0& 31.5& 26.1& 30.5 \\
LLaVA-Med-7B   & 25.3& 48.4& 35.9& 42.1& 45.2& 44.0& 31.7& 83.7& 44.5    \\
Qwen-VL-Chat & 51.5& 45.4& 43.9& 54.0& 55.4& 49.5& 63.1& 33.5& 49.5 \\
Yi-VL-34B & 39.8& 57.2& 51.4& 70.5& 54.5& 61.4& 64.2& 40.5& 54.9 \\
LLaVA-v1.6-7B   & 40.1& 39.5& 54.8& 58.4& 54.0& 48.8& 53.3& 47.9& 49.6  \\
LLaVA-v1.6-13B   & 40.0& 43.6& 47.4& 63.2& 58.0& 50.5& 59.6& 42.6& 50.6   \\
LLaVA-v1.6-34B   & 50.6& 63.4& 60.9& 68.4& 65.7& 62.8& 74.7& 44.5& 61.4  \\
\rowcolor{gray!17}\multicolumn{10}{l}{\textbf{Our Training}}                                     \\
LLaVA-v1.5-LLaMA3-8B  & 33.0 & 49.7 & 53.8 & 76.0 & 63.1 & 48.4 & 56.6 & 31.2 & 48.8   \\
\begin{tabular}[c]{@{}l@{}} \quad  +   LLaVA\_Med\end{tabular} & 60.8& 68.5& \textbf{66.3}& 79.0& 66.6& 60.3& 73.3& 49.3& 65.5 \\
\begin{tabular}[c]{@{}l@{}} \quad   +  \colorbox{lightblue}{PubMedVision}\end{tabular}  & \textbf{61.6} & \textbf{80.2} & 65.1& \textbf{86.3}& \textbf{71.6} & \textbf{67.4} & \textbf{81.4} & \textbf{87.4} & \textbf{75.1}   \\ \midrule
\textbf{HuatuoGPT-Vision-34B} & 60.8 & \textbf{85.5} & \textbf{66.5} & \textbf{90.0} & \textbf{74.0} & \textbf{71.3} & \textbf{83.8} & 81.7 & \textbf{76.7}
 \\ 
\bottomrule
\end{tabular}
\caption{\label{tb-res2} The accuracy of OmniMedVQA within different modalities. Specifically, \textbf{FP} denotes \textit{Fundus Photography}, \textbf{IRI} denotes \textit{Infrared Reflectance Imaging}, \textbf{MRI} denotes \textit{Magnetic Resonance Imaging}, \textbf{OCT} denotes \textit{Optical Coherence Tomography}, \textbf{Der} denotes \textit{Dermoscopy}, End denotes \textit{Endoscopy}, \textbf{Mic} denotes \textit{Microscopy Images}, \textbf{US}  denotes \textit{Ultrasound}.}
\end{table}

\paragraph{Traditional Medical Imaging Evaluation} OmniMedVQA integrates 41 traditional medical imaging tasks, all formatted as VQA. Table \ref{tb-res2} presents the results of it across 8 different modalities. After incorporating PubMedVision, the performance of LLaVA-v1.5-LLaMA3-8B showed a significant improvement of \textbf{26.3\%}, which is notably higher than the 16.7\% improvement achieved with the LLaVA\_Med dataset. With PubMedVision, LLaVA-v1.5-LLaMA3-8B outperforms previous open-source models.

% 测试集
\begin{table}[ht!]
\small \centering
\begin{tabular}{lccccc|c}\toprule
\textbf{Model}  &  \textbf{BMS} & \textbf{CM} & \textbf{DLM} & \textbf{P} & \textbf{PH} & \begin{tabular}[c]{@{}c@{}}MMMU\\ Health \& Medicine\end{tabular} \\ \midrule
Med-Flamingo &  29.6&  28.1&  24.8&  25.3&  31.2&  28.3  \\
RadFM  &  27.5&  26.8&  25.8&  24.7&  29.1&  27.0  \\
LLaVA-Med-7B &  39.9&  39.1&  34.6&  37.4&  34.0&  36.9 \\
Qwen-VL-Chat &  36.5&  31.7&  32.7&  28.4&  34.6&  32.7 \\
Yi-VL-34B &  49.4&  48.9&  43.2&  40.5&  32.0&  41.5 \\
LLaVA-v1.6-7B &  40.5&  36.9&  32.1&  32.3&  26.9&  33.1  \\
LLaVA-v1.6-13B  &  53.6&  46.7&  33.3&  22.2&  40.0&  39.3  \\
LLaVA-v1.6-34B  &  56.4&  56.0&  46.9&  \textbf{46.7}&  \textbf{41.7}&  48.8 \\
\rowcolor{gray!17}\multicolumn{7}{l}{\textbf{Our Training}}    \\
LLaVA-v1.5-LLaMA3-8B &  42.3&  44.0&  37.0&  34.7&  35.2&  38.2 \\ 
\begin{tabular}[c]{@{}l@{}}\quad +  LLaVA\_Med\end{tabular} &  48.2&  43.8&  42.0&  39.7&  35.8&  41.1 \\
\begin{tabular}[c]{@{}l@{}} \quad   +  \colorbox{lightblue}{PubMedVision}\end{tabular} &  \textbf{61.0}&  \textbf{58.8}&  \textbf{50.0}&  44.7&  38.7&  \textbf{49.1} \\ \midrule
\textbf{HuatuoGPT-Vision-34B} &  \textbf{64.6}&  \textbf{62.5}&  \textbf{50.6}&  \textbf{54.1}&  \textbf{44.2}&  \textbf{54.4} \\
\bottomrule
\end{tabular}
\caption{\label{tb-test_mmmu}Results on the \textbf{test set} for the MMMU Health \& Medicine track. The Health \& Medicine track is divided into five categories: \textbf{BMS} for \textit{Basic Medical Science}, \textbf{CM} for \textit{Clinical Medicine}, \textbf{DLM} for \textit{Diagnostics and Laboratory Medicine}, \textbf{P} for \textit{Pharmacy}, and \textbf{PH} for \textit{Public Health}. Results are obtained by submitting to the official website.}
\end{table}

\paragraph{MMMU Health \& Medicine Track} MMMU is a widely recognized multimodal benchmark, and we utilize its Health \& Medicine Track for assessment. Figure Table \ref{tb-test_mmmu} presents the results of the MMMU test set, showing that LLaVA-v1.5-LLaMA3-8B + PubMedVision surpassed other models in the Health \& Medicine Track, with performance comparable to the larger-parameter LLaVA-v1.6-34B. These findings further validate PubMedVision's effectiveness in aligning medical images. 

\begin{table}[ht!]
\centering \small
\begin{tabular}{lccccc} \toprule
 & \textbf{VQA-RAD} & \textbf{SLAKE} & \textbf{PathVQA} & \textbf{PMC-VQA}  \\ \midrule
\textbf{LLaVA-v1.5-7B}\cite{llava} & 50.6 & 53.4 & 52.3 & 33.1  \\
\quad  $\oplus$  \colorbox{lightblue}{PubMedVision}& 57.5\imp{6.9}     & 57.6\imp{4.3}   & 57.6\imp{4.3}   & 46.3\imp{13.2}  \\ \midrule
\textbf{Qwen-VL-Chat}\cite{qwen} & 47.0 & 56.0 & 55.1 & 36.6 \\
\quad  $\oplus$  \colorbox{lightblue}{PubMedVision} & 54.3\imp{7.3}    & 66.7\imp{10.7}  & 57.0\imp{1.9}    & 48.4\imp{11.8}    \\ 
\bottomrule
\end{tabular}
\caption{\label{tb-othermodel} PubMedVision for other MLLMs, where $\oplus$ denotes further training with PubMedVision.}
\end{table}

\paragraph{Applicability of PubMedVision} To verify the applicability of PubMedVision across different MLLM models, we further trained PubMedVision on other MLLM models, specifically LLaVA-v1.5-7B and Qwen-VL-Chat. As demonstrated in Table \ref{tb-othermodel}, PubMedVision effectively enhances the medical multimodal capabilities of these diverse MLLM models as well.

\subsection{Experiment 2: Data Quality of PubMedVision}
\label{section:ep2}

\paragraph{Experimental Setup} To validate the effect of the MLLM reformatter in PubMedVision, we constructed four datasets based on the four caption construction methods described in Section \ref{Section:2_2}. Specifically, we randomly sampled 60,000 image-context pairs from PubMedVision to create these four distinct datasets. For each caption, we pre-set the question: {\tt "Please provide a description of the given medical image"} to form VQA datasets, which we refer to as  \colorbox{lightpink}{Native-Captions-60K}, \colorbox{lightgreen}{LLM-Reformatted-60K}, \colorbox{lightyellow}{GPT4v-Distill-60K} and \colorbox{lightblue}{MLLM-Reformatted-60K}. Detailed explanations of these four methods are provided in Appendix \ref{ap-methoddetail}.

\begin{figure*}[ht!]
  \centering
  \resizebox{0.6\textwidth}{!}{
  \includegraphics[width=\textwidth]{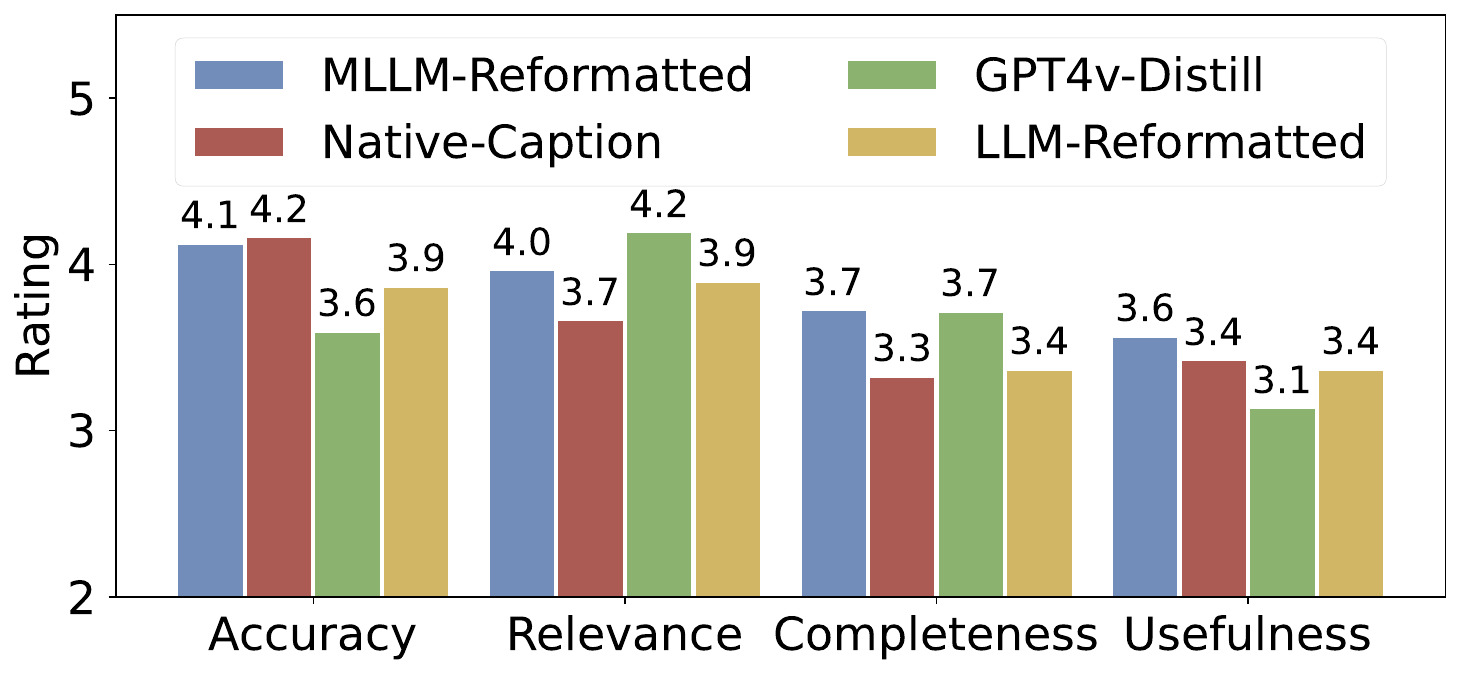}
  }
  \caption{\label{fig-rating}Scoring results from medical experts. Four scoring metrics are detailed in Appendix \ref{ap-scoringGuideline}.}
\end{figure*}

\paragraph{Expert Evaluation} To assess data quality, we randomly sampled 90 images, each contain 4 descriptions form \colorbox{lightpink}{Native-Captions-60K}, \colorbox{lightgreen}{LLM-Reformatted-60K}, \colorbox{lightyellow}{GPT4v-Distill-60K} and \colorbox{lightblue}{MLLM-Reformatted-60K}, totaling 360 entries. Three medical experts are invited to evaluate these image descriptions, each reviewing an equal number from each category. The criteria included: 1) \textbf{Accuracy}: correctness of the description, 2) \textbf{Relevance}: relevance to the image and avoidance of irrelevant details, 3) \textbf{Completeness}: inclusion of key medical features, and 4) \textbf{Usefulness}: utility for medical decision-making, diagnosis, and treatment planning. Each item is rated on a scale of 1-5. Detailed scoring criteria are in Appendix \ref{ap-scoringGuideline}. Table \ref{fig-rating} shows the scoring results (average values). 
Although \textit{Native-Captions} demonstrates high accuracy, it falls short in terms of relevance and completeness. \textit{LLM-Reformatted} shows improvements in relevance but remains deficient in completeness. \textit{GPT4v-Distill} excels in relevance and completeness, yet it underperforms in accuracy and usefulness. \textit{MLLM-Reformatted} excels across all metrics, offering the highest levels of completeness and usefulness along with substantial accuracy and relevance, indicative of superior overall quality.

\paragraph{Empirical Evaluation} Using LLaVA-v1.5-LLaMA3-8B, we evaluated four datasets to enhance medical multimodal capabilities. As shown in Figure \ref{tb-e2res2}, the \colorbox{lightblue}{MLLM-Reformatted} method outperforms other datasets with the same data volume, demonstrating superior alignment in medical multimodal applications. Additionally, a comparison between the full datasets of \colorbox{lightblue}{PubMedVision} and \colorbox{lightpink}{Native-Captions} reveals that PubMedVision performs significantly better, supporting the use of MLLMs for data reformatting.

\begin{table}[ht!]
\centering \small
\begin{tabular}{lccccc} \toprule
 & \textbf{VQA-RAD} & \textbf{SLAKE} & \textbf{PathVQA} & \textbf{PMC-VQA}  \\ \midrule
 \textbf{LLaVA-v1.5-LLaMA3-8B}              & 54.2    & 59.4  & 54.1    & 36.4   \\ 
\quad  +  \colorbox{lightpink}{Native-Caption-60K}  & 53.5    & 58.9  & 52.8    & 36.9   \\  
\quad  +  \colorbox{lightgreen}{LLM-Rephrase-60K}  & 56.5    & 63.7  & 54.0    & 39.1    \\ 
\quad  +  \colorbox{lightyellow}{GPT4v-Distill-60K} & 55.0    & 60.6  & 54.7   & 35.3   \\ 
\quad  +  \colorbox{lightblue}{PubMedVision-60K}& \textbf{56.8}    & \textbf{64.1}  & \textbf{55.1}   & \textbf{40.8}   \\ \cmidrule(r){2-5} 
\quad  +  Native Caption of \colorbox{lightblue}{PubMedVision} & 60.8    & 65.2  & 56.9    & 45.6  \\ 
\quad  +  \colorbox{lightblue}{PubMedVision} & \textbf{63.8}   & \textbf{74.5}  & \textbf{59.9}    & \textbf{52.7}  \\ 
\bottomrule
\end{tabular}
\caption{\label{tb-e2res2} Comparison of different datasets. The \textbf{60K} dataset is added only in the second stage of training. \textbf{Native Caption of PubMedVision} refers to using the original image captions, incorporated in both phases to match the training of \textbf{PubMedVision}.}
\end{table}

\section{Related Works}
\paragraph{Multimodal Large Language Models}  Recent advancements in MLLMs leverage the capabilities of LLMs such as LLaMA to integrate visual features into the textual space. Notably, Flamingo \cite{alayrac2022flamingo} introduces visual features by incorporating cross-attention layers into LLMs. To align multimodal features effectively, BLIP2 \cite{li2023blip} integrates a pre-trained visual encoder with LLMs through a novel Q-former. InstructBLIP \cite{dai2024instructblip} further refines this approach by enhancing performance using instruction-following data. Following this trend, LLaVA \cite{llava} and subsequent MLLMs \cite{zhu2023minigpt,ye2023mplug} utilize high-quality multimodal data for instruction tuning, demonstrating significant improvements. Additionally, ALLVA \cite{allava} shows that even a small model (3B) can achieve impressive results with high-quality Visual Question Answering (VQA) data. This underscores the importance of multimodal data.

\paragraph{Medical MLLMs}  Encouraged by the success of medical LLMs such as ChatDoctor \cite{yunxiang2023chatdoctor}, MedicalGPT \cite{MedicalGPT}, HuatuoGPT \cite{zhang2023huatuogpt, chen2023huatuogpt}, and Apollo \cite{wang2024apollo}, researchers have been focusing on developing a medical Multimodal LLM capable of understanding medical images. Med-Flamingo \cite{Med-flamingo} extends Flamingo to the medical domain by utilizing medical multimodal data for pre-training. LLaVA-Med \cite{llava-med-citation} refines this approach by filtering image-text pairs from PubMed papers and smaller VQA datasets synthesized by LLMs to train a medical MLLM based on LLaVA's parameters. Additionally, \cite{pmc-vqa} created the PMC-VQA dataset for medical VQA by self-instruction on PMC-OA \cite{pmc-oa}. Using this dataset, they developed MedVInT. RadFM \cite{RadFM} integrates a large amount of medical multimodal data, including 2D and 3D radiology images, to construct a radiology MLLM. However, according to recent findings \cite{omnimedvqa}, current medical models still lag behind general medical models in medical multimodal, indicating that higher quality datasets are needed for medical multimodal applications.

\paragraph{Medical VQA Datasets} To enhance image-text alignment and develop medical multimodal chatbots, researchers have focused on constructing medical VQA datasets. VQA-RAD \cite{rad-vqa}, SLAKE \cite{slake}, and Path-VQA \cite{pathvqa} are among the earliest medical VQA datasets. However, their sample sizes are small (less than 20K) and their diversity is limited, primarily to radiology modalities. Subsequently, PMC-VQA \cite{pmc-vqa} expands the dataset scale by using image-text data from PubMed papers and rewriting it into VQA format using LLMs. LLaVA-Med VQA \cite{llava-med-citation} data is derived from filtering higher quality data from PMC-15M  \cite{BiomedCLIP} and synthesizing VQA using LLMs. PMC-CaseReport \cite{rad-vqa} filters case images from PubMed and generates VQA using LLMs, though it retains only radiology modality images. Currently, there is still a need for more comprehensive and larger-scale medical VQA datasets.

\section{Conclusion}
In this study, we refined high-quality data from numerous medical image-text pairs on PubMed. We then employ MLLM-powered reformatting method to enhance this data. In this way, we construct PubMedVision, a large-scale, high-quality medical multimodal dataset. Experimental results show that PubMedVision significantly boosts the multimodal capabilities of MLLMs,  with marked improvements on benchmarks. This inspires the idea that PubMed holds great potential to advance medical multimodal capabilities, with the key challenge being how to improve data quality, despite the presence of many non-medical images and poor descriptions. We hope that the proposed PubMedVision dataset can aid the development of medical MLLMs in the future.

\bibliography{neurips_2024}
\bibliographystyle{unsrt}

%%%%%%%%%%%%%%%%%%%%%%%%%%%%%%%%%%%%%%%%%%%%%%%%%%%%%%%%%%%%
\clearpage

\appendix

% \section{Appendix / supplemental material}

\section{More Experiments}
\label{ap-moreexperiment}
\paragraph{Fine-tuned Results of VQA Benchmarks} To verify whether PubMedVision can enhance downstream tasks, we fine-tuned the model using the training set of the Benchmarks. As shown in Figure \ref{tb-finetuned}, PubMedVision effectively improves downstream medical tasks, significantly benefiting all four VQA downstream tasks.

% \definecolor{lightorange}{rgb}{1, 0.8, 0.6}
\definecolor{myorange}{RGB}{209, 133, 52}
\begin{table}[ht!]\centering\small
\begin{tabular}{lcccc|c}
\toprule
\textbf{Model} & \textbf{\begin{tabular}[c]{@{}c@{}}VQA-RAD\\ \textcolor{myorange}{(Finetuned)}\end{tabular}} & \textbf{\begin{tabular}[c]{@{}c@{}}SLAKE\\ \textcolor{myorange}{(Finetuned)}\end{tabular}} & \textbf{\begin{tabular}[c]{@{}c@{}}PathVQA\\ \textcolor{myorange}{(Finetuned)}\end{tabular}} & \textbf{\begin{tabular}[c]{@{}c@{}}PMC-VQA\\ \textcolor{myorange}{(Finetuned)}\end{tabular}} & \textbf{Avg.} \\ \midrule
\rowcolor{gray!17} \multicolumn{6}{l}{\textbf{Fine-tuning on the training set.}} \\
LLaVA-v1.5-LLaMA3-8B & 63.3 & 68.9 & 85.2 & 50.3 & 66.9 \\
\begin{tabular}[c]{@{}l@{}} \quad + LLaVA\_Med\end{tabular} & 66.3 & 69.5 & 90.7 & 52.7 & 69.8\\
\begin{tabular}[c]{@{}l@{}} \quad + \colorbox{lightblue}{PubMedVision}\end{tabular} & \textbf{68.9} & \textbf{84.1} & \textbf{93.0} & \textbf{57.3} &  \textbf{75.8} \\ \bottomrule
\end{tabular}
\caption{\label{tb-finetuned} Results on VQA Benchmarks after fine-tuning on the task training sets. All datasets were trained using their respective in-built training sets, over 2 training epochs.}
\end{table}

\paragraph{Results on validation set of MMMU} Table \ref{tb-val_mmmu} presents the validation results of MMMU, where LLaVA-v1.6-34B exhibits superior overall performance. However, compared to the test set results of MMMU (official submission) in Table \ref{tb-test_mmmu}, LLaVA-v1.5-LLaMA3-8B combined with PubMedVision demonstrates better performance. Overall, PubMedVision allows the 8B version of LLaVA to achieve effects comparable to the 34B version in medical applications.

\begin{table}[ht!]
\small \centering
\begin{tabular}{lccccc|c}\toprule
\textbf{Model}  &  \textbf{BMS} & \textbf{CM} & \textbf{DLM} & \textbf{P} & \textbf{PH} & \begin{tabular}[c]{@{}c@{}}MMMU\\ Health \& Medicine\end{tabular} \\ \midrule
Med-Flamingo  &  33.6&  30.2&  23.3&  29.3&  25.8&  28.4 \\
RadFM  &  31.6&  28.2&  26.7&  26.2&  26.8&  27.9 \\
LLaVA-Med-7B &  50.0&  33.3&  26.7&  40.7&  43.3&  38.6 \\
Qwen-VL-Chat &  39.3&  36.7&  20.0&  29.6&  33.3&  31.7 \\
Yi-VL-34B &  48.1&  55.6&  36.7&  48.1&  53.3&  48.2\\
LLaVA-v1.6-7B &  46.4&  33.3&  30.0&  29.6&  26.7&  33.1  \\
LLaVA-v1.6-13B  &  53.6&  46.7&  33.3&  22.2&  40.0&  39.3  \\
LLaVA-v1.6-34B  &  \textbf{57.1}&  \textbf{63.3}&  \textbf{50.0}&  44.4&  \textbf{63.3}&  \textbf{55.9} \\
\rowcolor{gray!17}\multicolumn{7}{l}{\textbf{Our Training}}    \\
LLaVA-v1.5-LLaMA3-8B   &  42.9&  43.3&  30.0&  25.9&  50.0&  38.6 \\ 
\begin{tabular}[c]{@{}l@{}} \quad +  LLaVA\_Med\end{tabular}  &  42.9&  46.7&  36.7&  40.7&  46.7&  42.8 \\
\begin{tabular}[c]{@{}l@{}}  \quad   +  \colorbox{lightblue}{PubMedVision}\end{tabular} &  50.0&  \textbf{63.3}&  36.7&  \textbf{48.1}&  53.3&  50.3 \\ \midrule
\textbf{HuatuoGPT-Vision-34B}  &  \textbf{64.3}&  60.0&  46.7&  \textbf{66.7}&  56.7&  \textbf{58.6}\\
\bottomrule

\end{tabular}
\caption{\label{tb-val_mmmu}Results on the \textbf{validation set} of MMMU Health \& Medicine track. The Health \& Medicine track is divided into five categories: \textbf{BMS} for \textit{Basic Medical Science}, \textbf{CM} for \textit{Clinical Medicine}, \textbf{DLM} for \textit{Diagnostics and Laboratory Medicine}, \textbf{P} for \textit{Pharmacy}, and \textbf{PH} for \textit{Public Health}.}
\end{table}

\section{Data Pipline}
\label{ap-datapipline}
To acquire a comprehensive dataset of PubMed images, we integrated previously compiled PubMed image and contextual text data, specifically LLaVA-Med PMC data (514K) \cite{llava-med-citation}, PMC-Inline (11M) \cite{rad-vqa}, and PMC-OA (1M) \cite{pmc-oa}. Although the dataset is extensive, most of the data consists of charts and graphs from papers rather than medical images. Therefore, we need to filter out higher-quality medical image-text data. We established a pipeline as follows: 

\begin{enumerate}
    \item \textbf{Contextual Text Filtering:}  Utilizing the SPECIALIST Lexicon \footnote{https://www.nlm.nih.gov/research/umls/new\_users/online\_learning/LEX\_001.html} from the Unified Medical Language System, we employed GPT-4 to filter out common phrases, creating a refined medical lexicon. Using this lexicon, we assessed the number of medical terms in image captions, filtering out data with fewer than five medical terms. This ensures the captions are sufficiently informative.

\item \textbf{Image Filtering:} Initially, we excluded images with a resolution lower than 336x336 pixels to ensure quality. Next, we filtered out chart images to retain only medical images. To accurately identify non-medical images, we manually labeled 1K images and synthesized 10K image labels using MLLMs (GPT4-Vision). We then trained a classifier based on the CLIP image encoder, achieving a 91\% accuracy on the validation set. This classifier is used to filter out non-medical images.

\item \textbf{Deduplication:}  We applied a semantic retriever for deduplication. Using all-mpnet-base-v2 \cite{reimers2019sentence} as the encoder, we generated semantic embeddings of the image captions. We then removed images with an embedding dot product similarity exceeding 480, ensuring a unique and high-quality dataset.

\end{enumerate}

\section{Question Set of Alignment VQA}
Alignment VQA is based on the generated image description \(d\) and the question \(q'\) sampled from a predefined question set. \(q'\) is sampled from the multi-image question set (Figure \ref{fig:vqa_multi_question}) if multiple images are involved, and from the single-image question set (Figure \ref{fig:vqa_single_question}) otherwise.

\label{ap-questionset}
\newtcolorbox{VQA_prompt}[1]{
        boxrule = 1.5pt,
        fontupper = \small\tt,
        fonttitle = \bf\color{black},
        arc = 5pt,
        rounded corners,
        colframe = black,
        colbacktitle = white!97!pink,
        colback = white!97!pink,
        title = #1,
}

\begin{figure}[ht!]
\begin{VQA_prompt}{Single-Image Question Set}
\begin{itemize}[leftmargin=3.5mm]
\setlength{\itemsep}{1.5pt}
    \item Please describe this picture.
    \item Can you describe the image for me?
    \item What details stand out in this image?
    \item Could you provide a detailed description of what is shown in the picture?
    \item What is the main focus of this photograph?
    \item Describe the composition and the subjects in this picture.
    \item Explain the visual content of the image
    \item Analyze the image in a comprehensive and detailed manner.
    \item Write a detailed description of the given image.
    \item What is this photo about?
    \item What is depicted in the image?
\end{itemize}
\end{VQA_prompt}
\caption{\label{fig:vqa_single_question} Single-image question set for alignment VQA. They convey the same meaning using different natural language expressions.}
\end{figure}

\begin{figure}[ht!]
\begin{VQA_prompt}{Multi-Image Question Set}
\begin{itemize}[leftmargin=3.5mm]
\setlength{\itemsep}{1.5pt}
    \item Please describe these pictures.
    \item Can you describe the images for me?
    \item What details stand out in these images?
    \item Could you provide a detailed description of what is shown in the pictures?
    \item What are the main focuses of these photographs?
    \item Describe the composition and the subjects in these pictures.
    \item Explain the visual content of the images.
    \item Analyze the images in a comprehensive and detailed manner.
    \item Write a detailed description of the given images.
    \item What are these photos about?
    \item What is depicted in the images?
\end{itemize}
\end{VQA_prompt}
\caption{\label{fig:vqa_multi_question} Multi-image question set for alignment VQA. They convey the same meaning using different natural language expressions.}
\end{figure}

\section{Prompts for different QA scenarios}
\label{ap-ScenariosPrompt}
\newtcolorbox{QA_prompt}[1]{
        boxrule = 1.5pt,
        fontupper = \small\tt,
        fonttitle = \bf\color{black},
        arc = 5pt,
        rounded corners,
        colframe = black,
        colbacktitle = white!97!black,
        colback = white!97!black,
        title = #1,
}

In our study, Instruction-Tuning VQA is generated based on ten pre-set different scenarios. This approach covers a broader range of medical topics and scenarios, thereby enhancing the diversity of the VQA pairs, and more comprehensively improving the ability to follow instructions. The sampling method also prevents the overconcentration or absence of certain scenarios, contributing to data balance, which in turn improves the performance and stability of the model.

\begin{figure}[H]
\begin{QA_prompt}{Standard Q\&A}
You need to generate a question-and-answer pair based on this image. The question should be designed to test other models' understanding of this medical image; it should be phrased simply and conversationally. However, your response should be professional, showcasing your understanding of the medical image by providing useful information derived from the image and detailed analysis. The reply should offer detailed and rich useful information.\\
\end{QA_prompt}
\caption{\label{QA-scenarios:standardQA} Prompt for Standard Q\&A Scenario: A guide for crafting a standard question-and-answer scenario.}
\end{figure}

\begin{figure}[H]
\begin{QA_prompt}{AI Model Assisting Doctor}
You need to generate a question-and-answer pair based on this image. You need to act as a doctor using an AI model to analyze a medical image to better understand a patient's condition. The doctor should ask specific questions about structures, abnormalities, and potential clinical significance visible on the image. The AI model should provide detailed analyses based on its algorithms but not make final clinical diagnoses. The doctor will use the information provided by the AI model to aid their diagnostic decision-making process.\\
\end{QA_prompt}
\caption{\label{QA-scenarios:ai-model-assisting-doctor} Prompt for AI Model Assisting Doctor Scenario: A simulated dialogue where a doctor consults an AI model about details in a medical image to improve diagnostic accuracy.}
\end{figure}

\begin{figure}[H]
\begin{QA_prompt}{AI Model Assisting Patient}
You need to generate a question-and-answer pair based on this image. You need to act as an AI model interacting with a patient who has questions about visible content on their medical image. The patient may be curious or confused about certain structures or markings on the image and seeks clear explanations. The AI model should explain specific details such as tissue density, shape, or any abnormal areas' potential meanings, maintaining simplicity and avoiding excessive medical jargon. The AI model's response should aim to provide educational information to help the patient better understand their imaging results, emphasizing that final interpretations and diagnoses must be done by a professional doctor.\\
\end{QA_prompt}
\caption{\label{QA-scenarios:ai-model-assisting-patient} Prompt for AI Model Assisting Patient Scenario: A simulated dialogue where an AI model explains details on a patient's medical image, aiming to clarify patient queries, while emphasizing that final interpretations are by professional doctors.}
\end{figure}

\begin{figure}[H]
\begin{QA_prompt}{Doctor and Patient's Family}
You need to generate a question-and-answer pair based on this image. You need to play the roles of a doctor and a patient's family member, discussing the results shown in the image. The doctor should explain the imaging findings in layman's terms and answer any questions posed by the family member. The family member may inquire about the cause of the disease, severity, treatment options, and related content. The doctor should answer patiently to ensure that the family member fully understands the condition.\\
\end{QA_prompt}
\caption{\label{QA-scenarios:doctor-and-patient-family} Prompt for Doctor and Patient's Family Scenario: A concerned family member inquiring about a patient's condition from the doctor.}
\end{figure}

\begin{figure}[H]
\begin{QA_prompt}{Doctor and Difficult Patient}
You need to generate a question-and-answer pair based on this image. You need to act as a doctor communicating with a patient who is skeptical about their diagnosis. The patient may pose a series of tricky questions, questioning the doctor's explanations and treatment suggestions. The doctor needs to use the imaging data patiently and explain the condition in an easy-to-understand manner, addressing all the patient's queries to alleviate their concerns and build trust.\\
\end{QA_prompt}
\caption{\label{QA-scenarios:doctor-and-difficult-patient} Prompt for Doctor and Difficult Patient Scenario: A simulated dialogue where a doctor patiently communicates a diagnosis to a skeptical patient, using the image data to explain the condition in a comprehensible way, and address all queries to build trust.}
\end{figure}

\begin{figure}[H]
\begin{QA_prompt}{Doctor to Doctor}
You need to generate a question-and-answer pair based on this image. This pair should be a professional discussion between doctors about the image. You need to mimic a doctor's tone in asking and answering questions. The response should provide detailed and rich useful information derived from the image.\\
\end{QA_prompt}
\caption{\label{QA-scenarios:doctor-to-doctor} Prompt for Doctor to Doctor Senario: A professional discussion scenario between doctors regarding a medical image.}
\end{figure}

\begin{figure}[H]
\begin{QA_prompt}{Evaluator and AI Model}
You need to generate a question-and-answer pair based on this image. You need to act as a member of a quality control team, focusing on assessing an AI model's visual capabilities in handling complex medical images. Team members should inquire about subtle details in the image.\\
\end{QA_prompt}
\caption{\label{QA-scenarios:evaluator-and-ai-model} Prompt for Evaluator and AI Model Scenario: A simulated interaction where a quality control team member assesses an AI model's ability to analyze complex medical images.}
\end{figure}

\begin{figure}[H]
\begin{QA_prompt}{Intern and Specialist Doctor}
You need to generate a question-and-answer pair based on this image. You should adopt the tone of an intern to ask questions and a specialist doctor to answer them. The answers should provide useful information derived from the image and give a detailed analysis. The response should provide detailed and rich useful information.
\\
\end{QA_prompt}
\caption{\label{QA-scenarios:intern-and-doctor} Prompt for Intern and Specialist Doctor Scenario: A simulated dialogue where an intern asks questions and a specialist provides detailed, informative answers based on a medical image.}
\end{figure}

\begin{figure}[H]
\begin{QA_prompt}{Medical Teacher and Student}
You need to generate a question-and-answer pair based on this image. You need to act as a medical teacher and a student, engaging in an educational interaction about the image. The teacher should pose questions, asking the student to analyze the image and propose possible diagnoses. The student should answer the questions and explain their observations and reasoning process.\\
\end{QA_prompt}
\caption{\label{QA-scenarios:teacher-and-student} Prompt for Medical Teacher and Student Scenario: A simulated educational interaction where the teacher prompts the student to analyze a medical image and propose potential diagnoses.}
\end{figure}

\begin{figure}[H]
\begin{QA_prompt}{Senior Doctor and Intern}
You need to generate a question-and-answer pair based on this image. You should act as a senior doctor and an intern, discussing the image. The senior doctor should pose relevant questions to test the intern's observational and analytical skills concerning the image, while the intern should respond and explain their viewpoint.\\
\end{QA_prompt}
\caption{\label{QA-scenarios:doctor-and-intern} Prompt for Senior Doctor and Intern Scenario: A simulated dialogue where a senior doctor tests an intern's observational and analytical skills through questions based on a medical image.}
\end{figure}

\section{Prompts for Evaluation}
\label{ap-evaluationTemplate}
\newtcolorbox{evaluation_prompt}[1]{
        boxrule = 1.5pt,
        fontupper = \small\tt,
        fonttitle = \bf\color{black},
        arc = 5pt,
        rounded corners,
        colframe = black,
        colbacktitle = white!97!blue,
        colback = white!97!blue,
        title = #1,
}

During the evaluation, we used a unified template.
\begin{figure}[H]
\begin{evaluation_prompt}{Prompt for Evaluation}
<question>\\
A. <option\_1>\\
B. <option\_2>\\
C. <option\_3>\\
D. <option\_4>\\
Answer with the option's letter from the given choices directly.\\
\end{evaluation_prompt}
\caption{\label{evaluation-prompt} Prompt for Evaluation.}
\end{figure}

\section{Comparison of Methods for Constructing Multimodal Datasets}
\label{ap-methoddetail}
Table \ref{tb-data2} presents four methods of synthesizing multimodal data. To facilitate a better comparison, we uniformly construct captions using these four methods. These captions are then combined with the query \textcolor{blue}{"Please provide a description of the given medical image"} to form a VQA dataset for comparing the differences among the various methods.

% \label{section:ep2}
\begin{table}[ht!]
\centering \small
\begin{tabular}{lp{0.7\linewidth}} \toprule
\textbf{Dataset} & \textbf{Caption Synthesis Method} \\ \midrule
 % \rowcolor{gray!17} \multicolumn{2}{l}{All datasets pre-set the question: \textcolor{blue}{"Please provide a description of the given medical image"}. } \\ \noalign{\smallskip}
\colorbox{lightpink}{Native-Caption-60K} & Uses the native contextual text (Caption and inline Mention) as the image caption. \\ \noalign{\smallskip}
\colorbox{lightgreen}{LLM-Reformatted-60K}   & Following the synthesis method of LLaVA-Med with LLMs, we provide only the contextual text to LLM (gpt-4-turbo-2024-04-09) to construct answers. For specific prompts, see Figure \ref{rephrase-prompt}. \\ \noalign{\smallskip}
\colorbox{lightyellow}{GPT4v-Distill-60K}  & We only provide the image to GPT-4-Vision (gpt-4-turbo-2024-04-09) to generate a description in response to the query \textcolor{blue}{"Please provide a description of the given medical image"}. \\ \noalign{\smallskip}
\colorbox{lightblue}{MLLM-Reformatted-60K} & The method of PubMedVision, where MLLMs construct data based on contextual text and visual information from the image. We use the answers from PubMedVision's Alignment VQA as the constructed caption. \\ 
\bottomrule
\end{tabular}
\caption{\label{tb-data2} Description of four methods for constructing image captions.}
\end{table}

\newtcolorbox{rephrase_prompt}[1]{
        boxrule = 1.5pt,
        fontupper = \small\tt,
        fonttitle = \bf\color{black},
        arc = 5pt,
        rounded corners,
        colframe = black,
        colbacktitle = white!97!blue,
        colback = white!97!blue,
        title = #1,
}

\begin{figure}[H]
\begin{rephrase_prompt}{Prompt for LLM-Reformatted}
You have been provided with textual context information of images from a biomedical research paper, but you do not have access to the actual image. You need to respond to the following question based on this image's context information. \\
\\
In your response, avoid using phrases like `mentioned', `caption', or `context'. Instead, describe the information as if it were directly observed `in the image'. Answer responsibly, avoiding any overconfidence, and refrain from giving medical advice or diagnostic information. Encourage the user to consult a healthcare professional for further advice.
\\
\\
<Image Context Information>: \colorbox{outerboxcolor}{\{image\_context\_information\}}
\\
\\
<Question>: Please provide a description of the given medical image.\\
\\
Please respond to the <Question> as instructed.
\end{rephrase_prompt}
\caption{\label{rephrase-prompt} Prompt for LLM-Reformatted. \colorbox{outerboxcolor}{\{image\_context\_information\}} pertains to image captions and inline mentions.}
\end{figure}

\section{Scoring Guidelines}
\label{ap-scoringGuideline}

\newtcolorbox{score_prompt}[1]{
        boxrule = 1.5pt,
        fontupper = \small\tt,
        fonttitle = \bf\color{black},
        arc = 5pt,
        rounded corners,
        colframe = black,
        colbacktitle = white!97!green,
        colback = white!97!green,
        title = #1,
}

\begin{figure}[H]
\begin{score_prompt}{Dataset Scoring Guidelines}
Please rate the image based on the following criteria (1 to 5):\\
\\
1. Accuracy: The rating should be based on whether the description accurately reflects the medical features and information in the image. A score of 5 indicates complete accuracy, while a score of 1 indicates complete inaccuracy.\\
\\
2. Relevance: The rating should be based on the degree of tight association between the described content and the key medical features and information in the image, and whether it successfully avoids the depiction of irrelevant details in the image. A score of 5 indicates high relevance, while a score of 1 indicates complete irrelevance.\\
\\
3. Completeness: The rating should be based on whether the description includes all the key medical features and information in the image. A score of 5 indicates complete completeness, while a score of 1 indicates significant omissions.\\
\\
4. Practicality: The rating should be based on the extent to which the description helps with medical decision-making, diagnosis, and treatment planning. A score of 5 indicates high practicality with in-depth descriptions, while a score of 1 indicates that the description is superficial and not practical.
\\
\end{score_prompt}
\caption{\label{dataset-scoring-guidelines} Dataset Scoring Guidelines.}
\end{figure}

\section{Limiations}
\label{ap-limitation}
The PubMedVision dataset has several limitations that should be considered:
\begin{itemize}
 \item \textbf{Hallucination of MLLMs}: The construction of the PubMedVision dataset utilizes MLLM models (GPT-4V), which as generative models, can produce hallucinations or inaccuracies. This might lead to errors in the dataset. Future studies may benefit from improved validation processes to mitigate this issue.
\item \textbf{Limited Scenario Diversity}: The Instruction-Tuning VQA of PubMedVision are  generated based on 10 predefined scenarios. This limited scope may have constrained the diversity of the dataset. Expanding the range of scenarios in future work could enhance the dataset's comprehensiveness and applicability to a wider array of medical situations.
\item \textbf{Data Selection}: The rigorous image selection strategy during data preparation ensured high-quality data but may have excluded potentially valuable data. Future data collection efforts could adopt a more balanced selection approach to optimize data utility.
\end{itemize}

\section{Ethical Statement}
\label{ap-EthicalStatement}
Our dataset was generated by the GPT4-V model, it may contain hallucinations or inaccuracies. Given this potential limitation, we strictly limit the use of the dataset to research purposes only. It is not to be employed in clinical or other industry applications where its use could lead to unintended consequences due to these possible inaccuracies. We emphasize the ethical responsibility of users to adhere to this restriction to ensure the safety and integrity of their applications.

\section{Case Study}
\label{ap-Casestudy}
\begin{table}
  \begin{minipage}{0.99\textwidth}
\centering  
\vspace{-4mm}
\scalebox{0.88}{
\begin{tabular}{l p{12.5cm} }
\toprule
 \multicolumn{2}{l}{\bf Sample 1: Standard Q\&A}  \\
\midrule
&  \includegraphics[height=3.5cm]{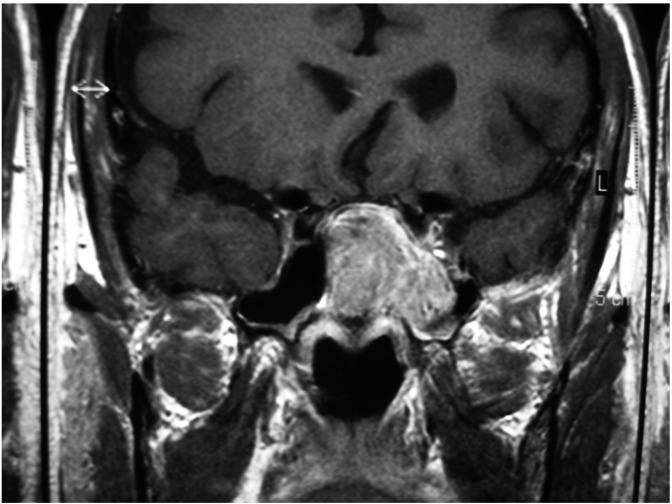} \\
\\
\textbf{Description} & The image shows a T1-weighted MRI scan with intravenous gadolinium enhancement. The scan reveals a well-defined, homogeneously enhancing mass located within the left sphenoid sinus. The mass appears to be occupying a significant portion of the sphenoid sinus, with no obvious signs of infiltration into adjacent structures. The surrounding soft tissues appear normal, and there is no evidence of bony erosion or destruction. Overall, the imaging findings are consistent with a sphenoid sinus mass, likely a benign tumor or other pathological process confined within the sphenoid sinus.
\\
\\
\textbf{QA-query} & What is the location of the mass observed in the MRI image?
\\
\\
\textbf{QA-answer} & The mass observed in the MRI image is located within the left sphenoid sinus. The sphenoid sinus is one of the paranasal sinuses, located deep within the skull base, behind the eyes. The MRI scan clearly shows the mass occupying a significant portion of the left sphenoid sinus, with no obvious extension into surrounding structures.
\\
\bottomrule
\end{tabular}
}
\vspace{1mm}
\caption{Sample 1 for Standard Q\&A Senorio.}
\label{sample 1}  
  \end{minipage}
\end{table}

\begin{table}
  \begin{minipage}{0.99\textwidth}
\centering  
\vspace{-4mm}
\scalebox{0.88}{
\begin{tabular}{l p{12.5cm} }
\toprule
 \multicolumn{2}{l}{\bf Sample 2: Evaluator and AI Model}  \\
\midrule
&  \includegraphics[height=3.5cm]{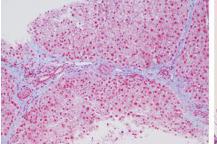} \\
\\
\textbf{Description} & This image depicts a microscopic view of a liver tissue section stained with trichrome, highlighting various histological features. The predominant colors are varying shades of pink and red, representing hepatocytes, and blue areas indicating connective tissue and collagen fibers. The structure of the tissue is organized with distinct lobular architecture. The hepatocytes are mostly uniform in size and appearance, with some regions showing slight signs of ballooning, indicative of early hepatocyte degeneration which is not overtly prominent. Sinusoids are visible between hepatocyte plates, and there are no clear signs of inflammation or fibrosis within the tissue section viewed.
\\
\\
\textbf{QA-query} & Given the subtle indications of hepatocyte changes in the image, can you identify any specific patterns or regions that may suggest early stages of ballooning degeneration?
\\
\\
\textbf{QA-answer} & Yes, some hepatocytes in the image show slight enlargement and pale cytoplasm which could be indicative of early stages of ballooning degeneration. These changes are subtle and not uniformly spread across the entire section, appearing more as isolated phenomena rather than widespread pathological changes.
\\
\bottomrule
\end{tabular}
}
\vspace{1mm}
\caption{Sample 2 for Evaluator and AI Model Senorio.} 
\label{sample 2}
  \end{minipage}
\end{table}

\begin{table}
  \begin{minipage}{0.99\textwidth}
\centering  
\vspace{-4mm}
\scalebox{0.88}{
\begin{tabular}{l p{12.5cm} }
\toprule
 \multicolumn{2}{l}{\bf Sample 3: Intern and Specialist Doctor}  \\
\midrule
&  \includegraphics[height=3.5cm]{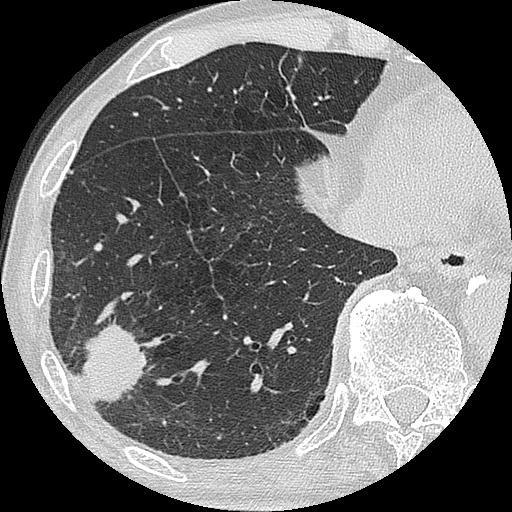} \\

\textbf{Description} & The image presents a cross-sectional view from a chest computed tomography (CT) scan, showcasing the right lung of the patient. Prominently visible is a solid mass with spiculated edges located in Segment 9b of the right lung. This mass exhibits a defined shape with a maximum diameter of approximately 30mm, indicating its noticeable size relative to surrounding lung structures. Additionally, the lung parenchyma displays scattered areas of decreased attenuation consistent with emphysematous changes, which are indicative of air trapping and lung tissue degradation.
\\
\\
\textbf{QA-query} & As an intern, I noticed a solid mass with spiculated edges on this CT scan. Could you explain the potential implications of these features and the significance of its location in the right lung's Segment 9b?
\\
\\
\textbf{QA-answer} & This CT scan reveals a solid mass with spiculated margins located in Segment 9b of the right lung, which significantly raises concerns for a malignant tumor, possibly a type of non-small cell lung cancer. Spiculated edges are typically suggestive of an aggressive tumor that may invade adjacent tissues, a finding that warrants further diagnostic evaluation and possibly biopsy to confirm malignancy. The location in Segment 9b, which is a part of the lung often less affected by other common conditions, emphasizes the need for targeted investigation. Managing such a case would likely involve a multidisciplinary approach including oncology, thoracic surgery, and radiology, to determine the best therapeutic strategies and follow-up care.
\\
\bottomrule
\end{tabular}
}
\vspace{1mm}
\caption{Sample 3 for Intern and Specialist Doctor Senorio.}
\label{sample 3}  
  \end{minipage}
\end{table}

\begin{table}
  \begin{minipage}{0.99\textwidth}
\centering  
\vspace{-4mm}
\scalebox{0.88}{
\begin{tabular}{l p{12.5cm} }
\toprule
 \multicolumn{2}{l}{\bf Sample 4: Doctor and Difficult Patient}  \\
\midrule
&  \includegraphics[height=3.5cm]{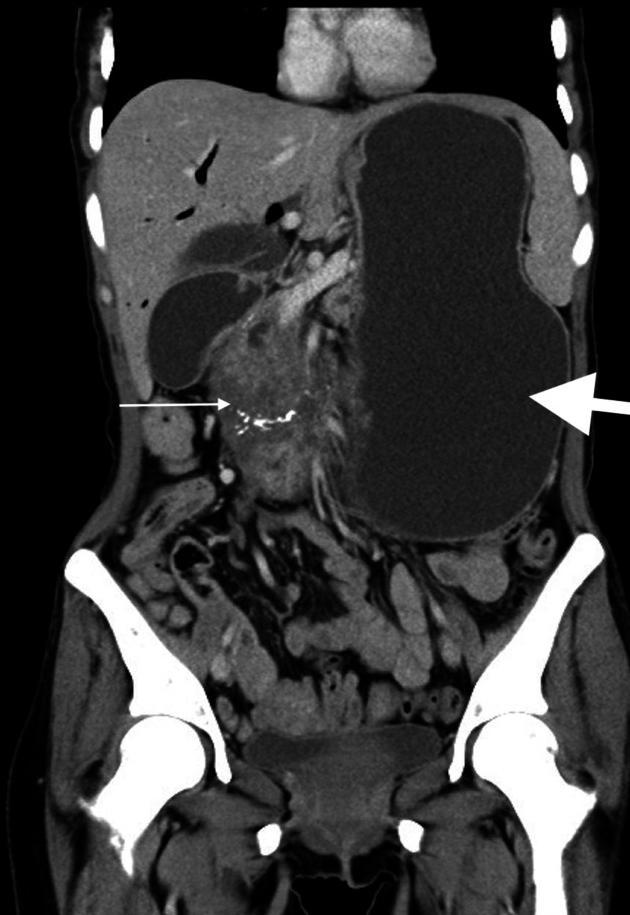} \\
\\
\textbf{Description} & The image is a coronal reformat of an abdominal CT scan. It distinctly displays two major findings. On the left side of the image (patient's right side), a mass is visible in the head of the pancreas, indicated by a thin arrow. This mass appears as a localized area of denser tissue compared to its surroundings. On the right side of the image (patient's left side), there is evidence of massive gastric dilatation, marked by a thick arrow. This dilatation shows the stomach expanded significantly, occupying a large portion of the abdominal cavity, which is abnormally large compared to the usual size of the stomach.
\\
\\
\textbf{QA-query} & Doctor, how serious is the mass in my pancreas and what are the potential risks with this gastric dilatation?
\\
\\
\textbf{QA-answer} & The mass in the head of your pancreas could represent several things ranging from benign conditions to more serious ones like pancreatic cancer. We will need further tests, possibly including a biopsy, to determine the exact nature of the mass. As for the gastric dilatation, it's quite significant and can lead to discomfort, pain, and other problems like nausea or vomiting if not managed timely. It may also indicate an obstruction or a severe motility disorder. Both conditions are serious and require careful evaluation and management. We'll plan some follow-up assessments to understand better and address both the mass and the dilatation appropriately, ensuring the best possible treatment plan is set for you.
\\
\bottomrule
\end{tabular}
}
\vspace{1mm}
\caption{Sample 4 for Doctor and Difficult Patient Senorio.}
\label{sample 4}  
  \end{minipage}
\end{table}

\begin{table}
  \begin{minipage}{0.99\textwidth}
\centering  
\vspace{-4mm}
\scalebox{0.88}{
\begin{tabular}{l p{12.5cm} }
\toprule
 \multicolumn{2}{l}{\bf Sample 5: Doctor and Patient’s Family}  \\
\midrule
&  \includegraphics[height=3.5cm]{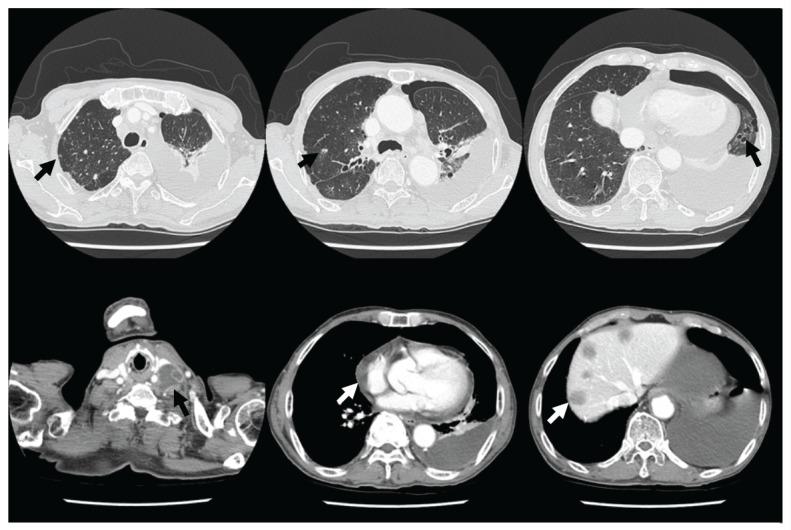} \\
\\
\textbf{Description} & The image displays a series of axial cuts of computed tomography (CT) scans from the thorax and abdomen. In the thoracic region, there are multiple tiny nodules and cystic lesions visible in both lungs. A significant left pleural effusion is evident with signs of pneumothorax, characterized by the presence of air in the pleural space which can be identified by the visceral pleural line and a clear space devoid of lung markings. Additionally, there is evidence of pericardial effusion as seen by the abnormal accumulation of fluid around the heart. The CT scans of the abdomen show enlargement of multiple lymph nodes and multiple hepatic masses, which appear as irregular areas within the liver.
\\
\\
\textbf{QA-query} & Doctor, can you explain what all these findings mean? How serious are these conditions and what are the possible treatments?
\\
\\
\textbf{QA-answer} & Certainly, I understand your concerns. The findings indicate a few serious conditions. Multiple tiny nodules and cystic lesions in the lungs could be indicative of an infectious or inflammatory process, or less likely, metastatic disease. The large amount of left pleural effusion and the presence of pneumothorax are concerning as they can compromise lung function. Pleural effusion means there is fluid accumulation that can restrict lung expansion, and pneumothorax involves air leaking into the space between the lung and chest wall, which can cause the lung to collapse. Pericardial effusion, the fluid around the heart, can affect heart function if severe. The enlarged lymph nodes and hepatic masses in the abdomen might suggest a possible malignancy or metastatic spread. Treatment options would vary based on the underlying cause but could include drainage of fluid from the pleural and pericardial spaces, antibiotics or chemotherapy if infection or malignancy are confirmed, and potentially surgery to address the hepatic masses. A multidisciplinary team approach will be crucial in managing these conditions effectively.
\\
\bottomrule
\end{tabular}
}
\vspace{1mm} 
\caption{Sample 5 for Doctor and Patient’s Family Senorio.}
\label{sample 5}  
  \end{minipage}
\end{table}

\begin{table}
  \begin{minipage}{0.99\textwidth}
\centering  
\vspace{-4mm}
\scalebox{0.88}{
\begin{tabular}{l p{12.5cm} }
\toprule
 \multicolumn{2}{l}{\bf Sample 6: Medical Teacher and Student (Multiple Images)}  \\
\midrule
&   
\begin{minipage}{.3\textwidth}
  \includegraphics[height=3.5cm]{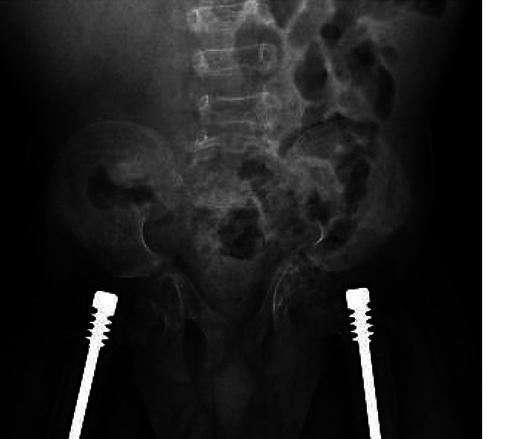}
\end{minipage}%
\begin{minipage}{.3\textwidth}
  \includegraphics[height=3.5cm]{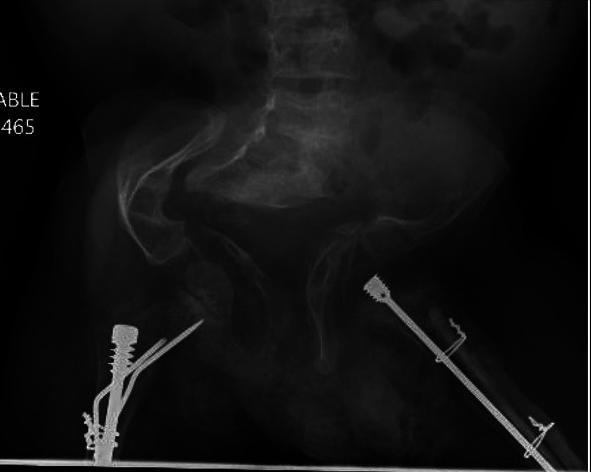}
\end{minipage}
\\
\\
\textbf{Description} & The provided images depict two AP (anteroposterior) pelvis radiographs. The first radiograph shows both hips with symmetric placement of screws or pins in what appears to be a bilateral surgical intervention, possibly for hip stabilization. There is visible deformation of the pelvic bones and the acetabulae (hip sockets) show signs of protrusion into the pelvic cavity. The second image features more pronounced skeletal changes including a notably misaligned right femur with surgical hardware adjustments, possibly indicating a post-injury or post-surgical state with a complex fixation device attached to the right side extending from the pelvis to the femur. These images suggest significant progression in the patient's condition, hinting at ongoing issues with hip stability and pelvic deformation.
\\
\\
\textbf{QA-query} & Given these pelvis radiographs, can you describe any observable changes or indicators that suggest a progression in the patient's pelvic condition?
\\
\\
\textbf{QA-answer} & Certainly, examining the radiographs, one can note an increased deformation in the pelvic region on the second image compared to the first. This includes greater protrusion of acetabular regions and a more complex assembly of surgical hardware on the right side, suggesting increased intervention. Such findings likely indicate a significant progression of underlying conditions affecting the pelvic structure which necessitated additional surgical measures over time.
\\
\bottomrule
\end{tabular}
}
\vspace{1mm} 
\caption{Sample 6 for Medical Teacher and Student Senorio. (Multiple Images)}
\label{sample 6}  
  \end{minipage}
\end{table}

\begin{table}
  \begin{minipage}{0.99\textwidth}
\centering  
\vspace{-4mm}
\scalebox{0.88}{
\begin{tabular}{l p{12.5cm} }
\toprule
 \multicolumn{2}{l}{\bf Sample 7: Evaluator and AI Model (Multiple Images)}  \\
\midrule
&   
\begin{minipage}{.23\textwidth}
  \includegraphics[height=3cm]{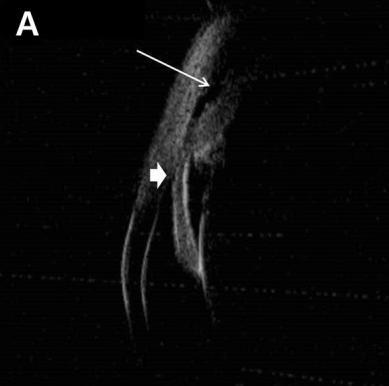}
\end{minipage}%
\begin{minipage}{.23\textwidth}
  \includegraphics[height=3cm]{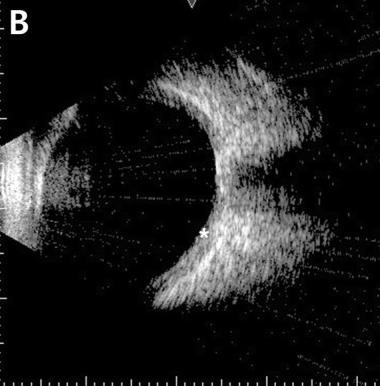}
\end{minipage}%
\begin{minipage}{.23\textwidth}
  \includegraphics[height=3cm]{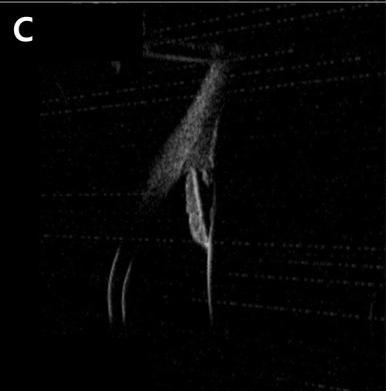}
\end{minipage}%
\begin{minipage}{.23\textwidth}
  \includegraphics[height=3cm]{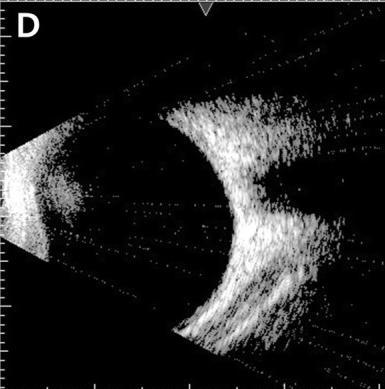}
\end{minipage}
\\
\\
\textbf{Description} & These images are medical ultrasound scans of the eye. Image A shows an ultrasound biomicroscopy (UBM) scan depicting angle closure with a shallow anterior chamber (indicated by the thick arrow) and anterior rotation of the ciliary process with ciliochoroidal effusion (thin arrow). Image B is a B-scan ultrasonography showing diffuse thickening of the choroid (asterisk). Image C, taken one month after treatment, shows a UBM scan of the left eye with a normal anterior chamber angle and the disappearance of ciliochoroidal effusion. Image D is another B-scan ultrasonography, which shows no choroidal thickening.
\\
\\
\textbf{QA-query} & In Image B, can you identify the specific location and extent of the choroidal thickening observed in the B-scan ultrasonography?
\\
\\
\textbf{QA-answer} & In Image B, the B-scan ultrasonography shows diffuse thickening of the choroid, as indicated by the asterisk. The thickening appears to be widespread throughout the choroidal region, extending across a significant portion of the eye's posterior segment.
\\
\bottomrule
\end{tabular}
}
\vspace{1mm} 
\caption{Sample 7 for Evaluator and AI Model Senorio. (Multiple Images)}
\label{sample 7}   
  \end{minipage}
\end{table}

\end{document}